\documentclass[10pt,twocolumn,letterpaper]{article}

\usepackage{cvpr}              

\definecolor{cvprblue}{rgb}{0.21,0.49,0.74}
\usepackage[pagebackref,breaklinks,colorlinks,allcolors=cvprblue]{hyperref}
\usepackage{array} 
\usepackage{booktabs}  
\usepackage{hhline}  
\usepackage{multirow} 
\usepackage{pifont}
\usepackage{hyperref}
\def\confName{CVPR}
\def\confYear{2025}

\title{EvaGaussians: Event Stream Assisted Gaussian Splatting from Blurry Images}

\author{
Wangbo Yu$^{1,2}$\textsuperscript{*}, Chaoran Feng$^{1}$\textsuperscript{*}, Jiye Tang$^{3}$, Jiashu Yang$^{4}$, Zhenyu Tang$^{1}$, Xu Jia$^{4}$, Yuchao Yang$^{1}$, \\
Li Yuan$^{1,2}$\textsuperscript{$\dag$} and Yonghong Tian$^{1,2}$\textsuperscript{$\dag$}\\ \\
$^1$Peking University, $^2$Peng Cheng Laboratory \\
$^3$University of Science and Technology of China \\
$^4$Dalian University of Technology \\
}

\begin{document}
\maketitle


\renewcommand{\thefootnote}{\fnsymbol{footnote}}
\footnotetext[1]{These authors contributed equally to this work.}
\footnotetext[2]{Corresponding author.}
\renewcommand{\thefootnote}{\arabic{footnote}}

\begin{abstract}
3D Gaussian Splatting (3D-GS) has demonstrated exceptional capabilities in synthesizing novel views of 3D scenes. However, its training is heavily reliant on high-quality images and precise camera poses. Meeting these criteria can be challenging in non-ideal real-world conditions, where motion-blurred images frequently occur due to high-speed camera movements or low-light environments.
To address these challenges, we introduce Event Stream Assisted Gaussian Splatting (\textbf{EvaGaussians}), a novel approach that harnesses event streams captured by event cameras to facilitate the learning of high-quality 3D-GS from blurred images. 
Capitalizing on the high temporal resolution and dynamic range offered by event streams, we seamlessly integrate them into the initialization and optimization of 3D-GS, thereby enhancing the acquisition of high-fidelity novel views with intricate texture details. 
To remedy the absence of evaluation benchmarks incorporating both event streams and RGB frames, we present two novel datasets comprising RGB frames, event streams, and corresponding camera parameters, featuring a wide variety of scenes and various camera motions. 
We then conduct a thorough evaluation of our method, comparing it with leading techniques on the provided benchmark. 
The comparison results reveal that our approach not only excels in generating high-fidelity novel views, but also offers faster training and inference speeds.
Video results are available at the \href{https://www.falcary.com/EvaGaussians/}{\textbf{project page}}.
\end{abstract}    
\vspace{-1.0em}
\section{Introduction}
\label{Introduction}
\begin{figure}[!ht]
    \begin{center}
    \includegraphics[width=1.0\linewidth,trim={0.2cm 0.8cm 0.2cm 0.8cm},clip]{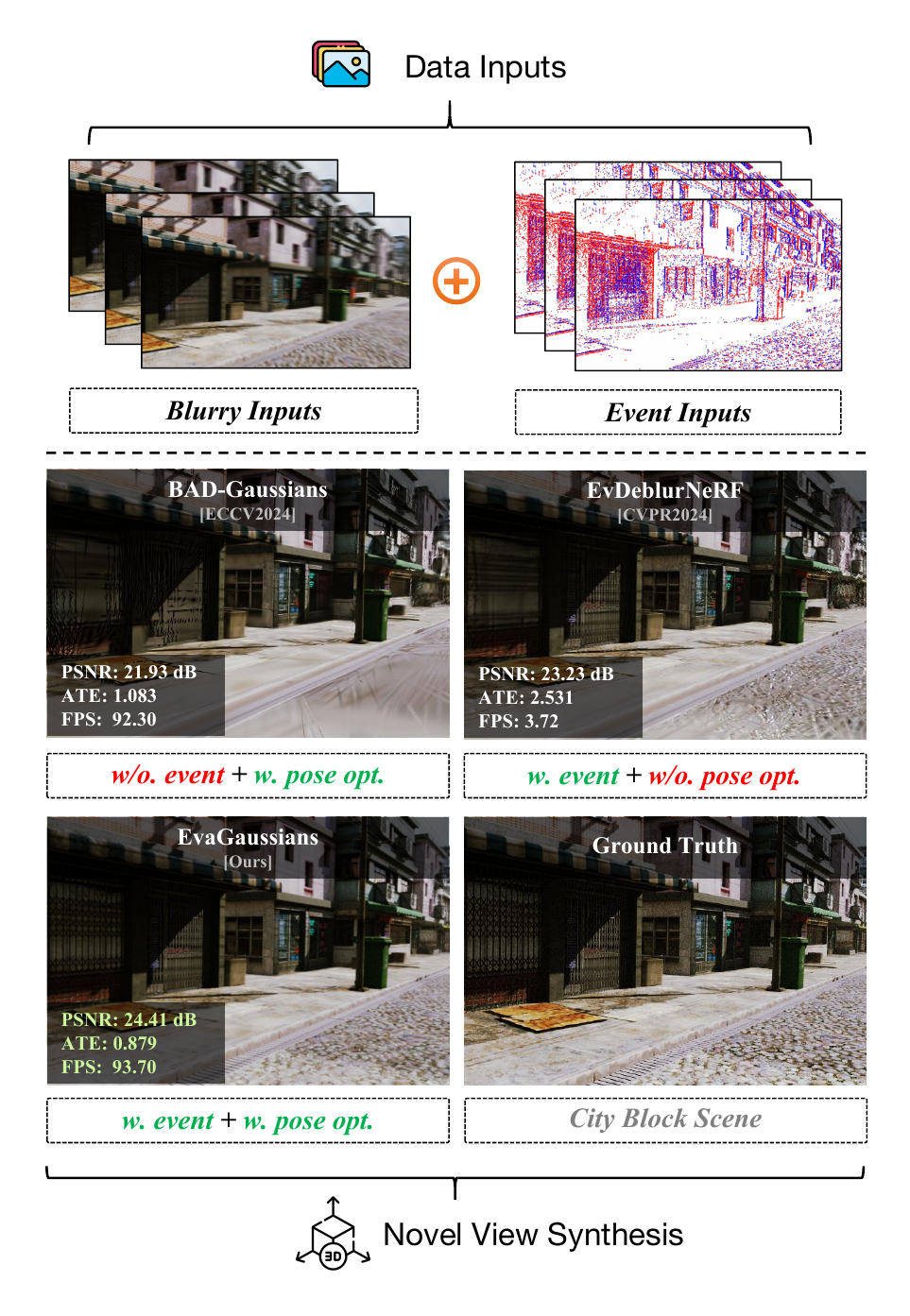}
    \end{center}
    \vspace{-1.0em}
    \label{pipeline}
    \caption{EvaGaussians integrates blurry images and event streams to reconstruct sharp 3D-GS for novel view synthesis.
    }
    \vspace{-0.5em}
    \label{fig:teaser}
\end{figure}
Novel view synthesis from 2D image collections has presented a persistent challenge within the field of computer vision and computer graphics. This task stands as a fundamental component in various vision applications, such as virtual reality~\cite{xu2023vr,laviola2008bringing,yu2024viewcrafter,tang2024cycle3d}, robotics navigation~\cite{rosinol2023nerfslam,zhu2022nice,yen2021inerf}, scene 
understanding~\cite{kerr2023lerf,liu2023weakly,liu2024openshape}, and many others, thereby prompting significant research efforts over the last decades. Amid pioneering works, 

3D Gaussian Splatting (3D-GS)~\cite{kerbl3Dgaussians} achieves notable success in generating high-fidelity novel views. It learns 3D Gaussians with the lightweight learnable parameters, and leverages a tile-based rasterization technique to render novel views, thereby surpassing NeRFs~\cite{nerf} in both training and rendering efficiency. 
However, the optimization of 3D-GS heavily relies on accurate camera poses and point cloud initialization produced by COLMAP~\cite{colmap}, which necessitates high-quality images without blurring and with adequate lighting. Fulfilling such conditions can be challenging in real-world situations. For example, in UAVs and robotics, rapid camera movement is common when capturing images or recording videos, which often result in significant motion blur. The mismatched features between blurred images can lead to inaccurate pose calibrations and point cloud initialization, thereby hindering the training process of 3D-GS. 

Recent studies have demonstrated the significant potential of event-based cameras in alleviating motion blur in images captured by conventional frame-based cameras~\cite{pan2019edi,jiang2020emd,lin2020edv,tangzhenyu2024cycle3d,yuanshenghai2024chronomagic,jinpeng2024moh,tang2024cycle3d}. Serving as an innovative bio-inspired visual sensor, event cameras asynchronously report the logarithmic intensity changes of each pixel captured, and can record higher temporal resolution and dynamic range data in contrast to conventional cameras. Motivated by this, prior works~\cite{qi2023e2nerf,cannici2024mitigating}, have attempted to leverage the event streams captured by event cameras to supervise the training of NeRFs. However, achieving real-time rendering and synthesizing high-fidelity novel views with intricate details poses substantial challenges for these methods.

To address these challenges, we introduce Event Stream Assisted Gaussian Splatting (\textbf{EvaGaussians}), which leverages the event streams captured by event cameras to enhance the learning of high-quality 3D-GS from motion-blurred images.
Harnessing the exceptional temporal resolution and dynamic range offered by event streams, we use them to assist in the initialization of 3D-GS, and incorporate them to jointly optimize 3D-GS and camera trajectories of blurry images through a blur reconstruction loss and an event reconstruction loss.
Due to the geometric ambiguity caused by blurry images, we further propose two event-assisted depth regularization terms to stabilize the geometry of 3D-GS. Through optimizing the 3D-GS in a progressive manner, our method can recover a high-quality 3D-GS that facilitates the real-time generation of high-fidelity novel views.
To summarize, our contributions can be delineated as follows:
\begin{itemize}

\item
We propose Event Stream Assisted Gaussian Splatting (\textbf{EvaGaussians}), a framework tailored for reconstructing a high-quality 3D-GS from motion-blurred images with the assistance of event camera. Once trained, our method is capable of recovering intricate details of the input blurry images and allows high-fidelity real-time novel view synthesis.

\item 
We contribute two novel datasets, including a synthetic dataset containing diverse scenes with various scales, and a real-world dataset captured by the color DAVIS346 event camera~\cite{davis346}, both feature various camera motions.  We believe they will set a benchmark for future researches.

\item  
We conduct a comprehensive evaluation of the proposed method and compare it with several strong baselines. The results reveal that our approach not only excels in generating high-fidelity novel views but also provides faster training and inference speeds.
\end{itemize}

\section{Related Works} 
\label{sec:related_works}



\subsection{Reconstructing 3D Scene from Blurry Images}

Reconstructing a high-quality 3D Scene typically requires high-fidelity, sharp images as supervision. However, motion-blurred images often occur in real world scenarios, thus hindering accurate reconstruction of 3D scenes. 
Several studies have been proposed to address this issue. For example, Deblur-NeRF~\cite{deblur-nerf} and DP-NeRF~\cite{lee2023dp} attempted to learn a blur formation kernel to model the image blurring process. BAD-NeRF~\cite{wang2023badnerf} further physically modeled the blurry images formation process, and adopted a bundle-adjustment strategy to jointly optimize NeRF parameters and the camera poses during the exposure time. 
These NeRF-based methods lacked real-time rendering capabilities and suffered from extended training times. With the rapid advancement of 3D-GS, a concurrent work, BAD-Gaussians~\cite{zhao2024badgs}, proposed to utilize 3D-GS as representation and follow the blur modeling and bundle-adjustment strategy adopted in~\cite{wang2023badnerf} to achieve deblurring reconstruction.
Although it achieved real-time rendering and faster convergence compared with prior works, it still struggled to handle severely blurred images in which COLMAP~\cite{colmap} will fail to produce the initial point clouds. Furthermore, it employed linear interpolation between the start and end camera poses to model camera trajectory during exposure time, necessitating careful selection of poses for more stable optimization. 

\subsection{Reconstructing 3D Scene from Event Streams}
Motivated by the exceptional properties offered by event cameras, several studies attempted to reconstruct 3D scenes from event streams captured by event cameras, particularly in low-light conditions with fast camera motion. 
For example, EventNeRF~\cite{rudnev2023eventnerf}, Ev-NeRF~\cite{hwang2023ev} and other concurrent works~\cite{wu2024ev-gs,xiong2024event3dgs,yin2024e-3dgs,zhang2024elite-ev-gs,yu2024viewcrafter} explored the reconstruction of a 3D representation from a rapidly moving event camera. 
Robust e-NeRF~\cite{low2023robust} and its variants~\cite{low2025deblur-e-nerf} further extended this task to the more challenging scenario of non-uniform camera motion, taking into account the refractory period of event cameras. 
These methods were typically designed to be supervised solely by information captured from a single event camera.
Recently, E-NeRF~\cite{enerf}, \({\textnormal{E}^2}\)\textnormal{NeRF}~\cite{qi2023e2nerf}, and EvDeblurNeRF~\cite{evdeblurnerf} proposed to jointly utilize event streams captured by event cameras and motion-blurred images captured by standard frame-based cameras to reconstruct a NeRF representation. 
Compared to methods that rely solely on event cameras, these methods can recover accurate color details. 
Additionally, in contrast to RGB-only methods, they are better at handling motion blur.
However, these NeRF-based methods suffer from long training and inference times, and face instability during training, which limit their further application.

\begin{figure*}[t]
    \begin{center}
    \includegraphics[width=1\linewidth,trim={0.0cm 0.0cm 0.0cm 0.0cm},clip]{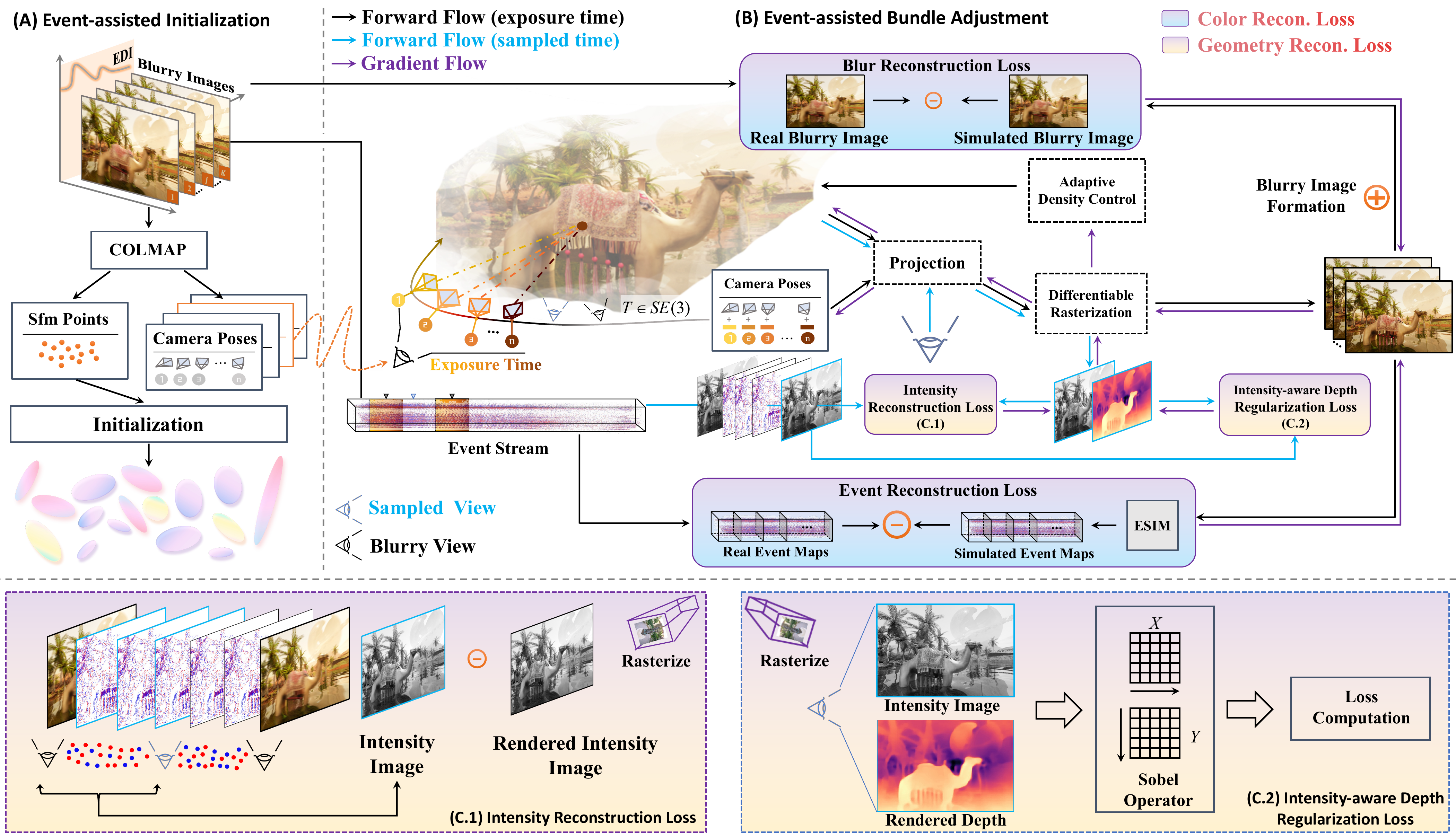}
    \end{center}
    \vspace{-0.9em}
    \caption{\textbf{Overview of EvaGaussians}. We use event streams to assist in the initialization of 3D-GS and incorporate them to jointly optimize both 3D-GS and the camera trajectories of blurry images during the exposure time, utilizing a blur reconstruction loss and an event reconstruction loss. Additionally, we propose two event-assisted depth regularization terms to stabilize the geometry of 3D-GS.
    }
    \label{fig:overview}
\end{figure*}
\section{Method}
\label{sec:method}

\subsection{Preliminary}

Event camera is a type of bio-inspired sensor that can asynchronously record intensity changes~\cite{gallego2020survey}.  In contrast to conventional cameras that are restricted to sequentially produce frames at a fixed frame rate, event cameras asynchronously trigger events in each pixel when their intensity change exceeds a constant threshold, featuring properties such as low latency and high dynamic range. Formally, let $\mathbf{I}_{xy}(t)$ denote the instantaneous intensity at pixel coordinate $(x, y)$ at time $t$, and $\mathbf{L}_{xy}(t)$ denotes its logarithm. An event $p = \pm 1$ will be triggered whenever the change of $\mathbf{L}_{xy}(t)$ surpasses the threshold $c$, where the polarity represents the direction (increase or decrease) of changes. Let $\delta_{t_0}(t)$ be the impulse function at time $t_0$ with a unit integral, the event can therefore be expressed as a continuous-time signal $\mathbf{e}_{xy}(t) = p\, \delta_{t_0}(t)$, where $t_0$ signifies the time at which the event occurs. Then, the proportional intensity change during a time interval $[s, t]$ can be computed as the integral of events that occurred between times $s$ and $t$, expressed as
$\mathbf{E}_{xy}(t) = \int_{s}^t \mathbf{e}_{xy}(h) dh$.
Given that each pixel can be treated separately in the event camera, the subscripts can be omitted:
\begin{equation} 
\label{eq:event} 
\begin{split}
\mathbf{E}(t) &= \int_{s}^t \mathbf{e}(h) dh.\\
\end{split}
\end{equation}
We can then represent the logarithmic intensity change as: $\mathbf{L}(t) - \mathbf{L}(s) = c\,\mathbf{E}(t)$, rewrite as $\mathbf{L}(t) = \mathbf{L}(s) + c\,\mathbf{E}(t)$, and subsequently obtain the actual intensity change:
\begin{equation}    
\label{eq:2-image}
\mathbf{I}(t) = \mathbf{I}(s) \cdot \exp( c\,  \mathbf{E}(t)).
\end{equation}
Therefore, when an image \(\mathbf{I}(s)\) is captured at time \(s\), and the event stream is recorded during the time interval \([s, t]\), the image \(\mathbf{I}(t)\) can be obtained by warping $\mathbf{I}(s)$ using Eq.~\ref{eq:2-image}.

\subsection{Event-assisted Initialization}
The optimization of 3D-GS requires camera calibration and point cloud initialization using COLMAP~\cite{colmap}. However, this process can fail when dealing with images that have significant motion blur.
Motion-blurred images are resulted from camera movements during the exposure time, which can be mathematically represented as:
\begin{equation}
\label{eq:blur}
\mathbf{B} =  \frac{1}{\tau} \int_{s-\tau/2}^{s+\tau/2} \mathbf{I}(t) dt ,
\end{equation}
where $\mathbf{B}$ denotes a captured blurry image, which is equivalent to averaging the instantaneous latent images $\mathbf{I}(t)$ during the exposure time $[s-\tau/2, s+\tau/2]$. 

To obtain initial camera poses and point clouds for 3D-GS optimization, we first preprocess the motion-blurred images using the Event-based Double Integral (EDI)~\cite{pan2019edi} model, which can be derived through substituting Eq.~\ref{eq:2-image} into Eq.~\ref{eq:blur}:
\begin{equation}
\label{eq:edi}
\mathbf{B} = \mathbf{I}(s) \cdot  \frac{1}{\tau} \int_{s-\tau/2}^{s+\tau/2} \exp( c\,  \mathbf{E}(t)) dt.
\end{equation}
Given the predefined threshold $c$, a blurry image $\mathbf{B}$, and the recorded event stream $\mathbf{E}(t)$, the EDI model (Eq.~\ref{eq:edi}) allows the derivation of $\mathbf{I}(s)$, following which the latent image $\mathbf{I}(t)$ at any moment within the exposure time can be estimated through Eq.~\ref{eq:2-image}. 
As shown in Figure.~\ref{fig:overview}(A), given a total of $K$ blurry images $\{\mathbf{B}^j\}_{j=1}^{K}$,
for each of them, we uniformly sample $n$ time stamps during their exposure time to obtain a series of EDI-estimated latent images rich in texture features, denoted as $\{\mathbf{I}_i\}_{i=1}^{n}$, then obtain their poses $\{\mathbf{P}_i\}_{i=1}^{n}$ and the initial point cloud of the scene using COLMAP~\cite{colmap}. 

After initialization, a straightforward approach to optimizing the 3D-GS is to use the EDI-estimated latent images and poses as supervision. However, although these images provide more texture features than the original blurry image, they still do not fully recover the ideal latent image and exhibit relatively low visual quality, which also introduces inaccuracies into the camera poses, thereby leading to unsatisfactory optimization results.
To more robustly recover a sharp 3D-GS from motion-blurred images, we propose to harness the advantages of event streams and seamlessly integrate them into the optimization process of 3D-GS.

\subsection{Event-assisted Bundle Adjustment}
As introduced in Eq.~\ref{eq:blur}, during the exposure time, a motion-blurred image can be decomposed into a series of latent images along a specific camera trajectory, which can be roughly approximated by the EDI-produced camera poses $\{\mathbf{P}_i\}_{i=1}^{n}$ according to Eq.~\ref{eq:edi}. Motivated by this, we jointly optimize these camera poses and the 3D-GS attributes in a bundle adjustment manner~\cite{wang2023badnerf} to simultaneously recover the blur-formation camera trajectories and a sharp 3D-GS. 
As shown in Figure.~\ref{fig:overview}(B), we add each of the EDI-produced camera poses a learnable offset $\{\mathbf{d}_i\}_{i=1}^{n}$ as correction parameters, resulting a learnable camera trajectory $\{\mathbf{\widetilde{P}}_i\}_{i=1}^{n}$, where $\mathbf{\widetilde{P}}_i = \mathbf{P}_i + \mathbf{d}_i$. 
In each training iteration, we simultaneously render $n$ images $\{\mathbf{\widetilde{I}}_i\}_{i=1}^{n}$ from the 3D-GS along the camera trajectory of the blurry view, and simulate the formation of motion-blurred images using a discrete approximation of Eq~\ref{eq:blur}, expressed as $\mathbf{\widetilde{B}}  = \frac{1}{n} \sum_{i=1}^{n} \mathbf{\widetilde{I}}_i$.
Consequently, for a total of $K$ real-captured blurry images $\{\mathbf{B}^j\}_{j=1}^{K}$, we can obtain their simulated versions $\{\mathbf{\widetilde{B}}^j\}_{j=1}^{K}$ through each corresponding learnable camera trajectory.

\noindent\textbf{Blur Reconstruction Loss.} With the simulated blurry images, we use the real captured blurry images $\{\mathbf{B}^j\}_{j=1}^{K}$ to serve as image level supervision. Specifically, for each blurry image \(\mathbf{B}^j\) and its simulated version \(\mathbf{\widetilde{B}}^j\), we employ a blur reconstruction loss to minimize their photometric error, expressed as
\begin{equation}
	\mathcal{L}_{blur} = (1-\lambda_1) \cdot \|\mathbf{B}^j - \mathbf{\widetilde{B}}^j\|_{1} + \lambda_1 \cdot \text{D-SSIM}(\mathbf{B}^j,\mathbf{\widetilde{B}}^j).
\end{equation} 
The formulation of blur reconstruction loss is the same as in the original 3D-GS~\cite{kerbl3Dgaussians}, it differs in utilizing blurry images as supervision and jointly optimizing the 3D-GS attributes and the camera trajectories, thus facilitating an initial deblurring reconstruction of 3D-GS.

\noindent\textbf{Event Reconstruction Loss.} 
Leveraging the abundant high-frequency information offered by the event streams, we further adopt an event reconstruction loss to aid in 3D-GS optimization.
Specifically, we uniformly divide the exposure time into $m = n - 1$ intervals, each with a duration of $\frac{\tau}{m}$. Subsequently, we integrate the recorded event stream along these time intervals using Eq.~\ref{eq:event}, resulting in $m$ event maps $\{\mathbf{E}_i\}_{i=1}^{m}$ to serve as event level supervision. During training, for the $j$-th blurry view, we convert the rendered image sequence $\{\mathbf{\widetilde{I}}_i\}_{i=1}^{n}$ on the camera trajectory into event maps $\{\mathbf{\widetilde{E}}_i\}_{i=1}^{m}$, using a differentiable event simulator~\cite{esim,vid2e}, and constrain the discrepancies between the simulated event maps and the ground truth event maps, expressed as:
\begin{equation}
\label{eq:6-event_loss}
	\mathcal{L}_{event} = \frac{1}{m} \sum_{i=1}^{m} \|\mathbf{E}_i - \mathbf{\widetilde{E}}_i\|_{1}.
\end{equation} 
The event reconstruction loss further aids in recovering a sharp 3D-GS with improved texture details.
\subsection{Event-assisted Geometry Regularization}
The blurry color images are captured only during the exposure time and are much sparser than the event stream. Relying on such low-quality image-level supervision may cause the 3D-GS to overfit on the training images, resulting in significant floaters and inferior geometry, which affects the quality of novel view synthesis.
Leveraging Eq.~\ref{eq:event} and Eq.~\ref{eq:edi}, given the continuously recorded event streams $\mathbf{E}(t)$, we can derive continuous grayscale intensity images $\mathbf{G}(t)$ that are rich in geometric information and can function beyond the exposure time. Motivated by this, we further propose two event-assisted geometry regularization terms to aid in 3D-GS training.

\noindent\textbf{Intensity Reconstruction Loss.} 
As shown in Figure.~\ref{fig:overview}(C.1), during training, we randomly sample continuous time $t$ between the interval of two adjacent blurry image, and derive the grayscale intensity image $\mathbf{G}(t)$ using Eq.~\ref{eq:2-image}. We then minimize the difference between it and the rendered intensity image from 3D-GS, expressed as:
\begin{equation}
	\mathcal{L}_{int} = (1-\lambda_2) \cdot \|\mathbf{G}(t) - \mathbf{\widetilde{G}}(t)\|_{1} + \lambda_2 \cdot \text{D-SSIM}(\mathbf{G}(t),\mathbf{\widetilde{G}}(t)),
\end{equation} 
where $\mathbf{\widetilde{G}}(t)$ is converted from the colored render result. 

\noindent\textbf{Intensity-aware Depth Regularization Loss.} 
As shown in Figure.~\ref{fig:overview}(C.2), to further improve the geometry of 3D-GS, inspired by~\cite{heise2013pm,comi2024snap}, we adopt an intensity-aware depth regularization loss during training, defined as:
\begin{equation}
\label{eq:depth}
\begin{aligned}
\mathcal{L}_{depth} = & \frac{1}{N} \sum_{x, y} (~|\partial_x \mathbf{\widetilde{D}}_{xy}(t)| e^{-\beta |\partial_x \mathbf{G}_{xy}(t)|}  \\
& + |\partial_y \mathbf{\widetilde{D}}_{xy}(t)| e^{-\beta |\partial_y \mathbf{G}_{xy}(t)|}~),
\end{aligned}
\end{equation}
where $\mathbf{\widetilde{D}}(t)$ is the rendered depth map, $(x,y)$ denotes the pixel location, $N$ is the total number of pixels, and $\beta$ is set to $2$ in our experiments. The horizontal and vertical gradients are calculated by applying convolution operations with $5\times5$ Sobel kernels~\cite{vairalkar2012edge-sobel}.
This regularization is founded on the observation that depth transitions in an image often correspond to changes in intensity. 
Therefore, it ensures that the spatial variation of depth closely matches that of the intensity image, thereby reducing geometric artifacts at object boundaries. 

\begin{table*}[!ht]
    \centering
    \caption{Quantitative comparisons of novel view synthesis across large-scale, medium-scale, object-level, and real-world scenes. The table reports the average performance for each scale, demonstrating that our method consistently surpasses previous state-of-the-art approaches across all metrics. Best-performing results are highlighted in \textbf{bold} and second-best results in \underline{underline}.}
    \label{tab:quan_table}
    \renewcommand{\arraystretch}{1.35}
    \resizebox{\linewidth}{!}{
        \begin{tabular}{c||c||c c | c c c| c c c c||>{\centering\arraybackslash}p{1.5cm}}
            \toprule
                \textbf{Scene Type} & \textbf{Metric} & \textbf{B-NeRF} & \textbf{B-3DGS} & \textbf{UFP-GS} & \textbf{EDI-GS} & \textbf{EFN-GS} & \(\mathbf{E}^2\)\textbf{NeRF} & \textbf{BAD-NeRF} & \textbf{BAD-GS} & \textbf{EDNeRF} & \textbf{Ours}\\
            \hhline{=::=::=========::=}
                \multirow{3}{*}{\textbf{Large-scale}}
                & {PSNR$\uparrow$}    & 21.33 & 21.48 & 21.36 & 22.31 & 22.69 & 22.96 & 23.85 & 23.86 & \underline{24.63} & \textbf{26.02} \\
                & {SSIM$\uparrow$}    & .6781 & .6876 & .6600 & .6855 & .6826 & .7066 & .7323 & .7325 & \underline{.7525} & \textbf{.8064} \\
                & {LPIPS$\downarrow$} & .4249 & .3971 & .3736 & .3823 & .3631 & .3751 & .3480 & .3473 & \underline{.3279} & \textbf{.2680} \\
            \hhline{-||-||---------||-}
                \multirow{3}{*}{\textbf{Medium-scale}}
                & {PSNR$\uparrow$}    & 24.08 & 24.80 & 26.38 & 26.44 & 26.13 & 27.78 & 28.46 & 28.46 & \underline{28.91} & \textbf{30.47} \\
                & {SSIM$\uparrow$}    & .7173 & .7512 & .8022 & .8012 & .7981 & .8656 & .8791 & .8789 & \underline{.8854} & \textbf{.9164} \\
                & {LPIPS$\downarrow$} & .3617 & .3187 & .2639 & .2581 & .2726 & .1985 & .1823 & .1816 & \underline{.1692} & \textbf{.1519} \\
            \hhline{-||-||---------||-}
                \multirow{3}{*}{\textbf{Objects}}
                & {PSNR$\uparrow$}    & 22.28 & 22.34 & 25.16 & 24.94 & 25.45 & 29.61 & 27.33 & 27.86 & \underline{29.83} & \textbf{30.24} \\
                & {SSIM$\uparrow$}    & .9041 & .9049 & .9275 & .9248 & .9289 & .9638 & .9476 & .9501 & \underline{.9655} & \textbf{.9698} \\
                & {LPIPS$\downarrow$} & .1479 & .1471 & .1174 & .1208 & .1103 & .0735 & .0928 & .0911 & \underline{.0722} & \textbf{.0702} \\
            \hhline{=::=::=========::=}
                \multirow{5}{*}{\textbf{Real-world}}
                & {BRISQUE$\downarrow$} & 92.25 & 73.80 & 62.94 & 62.75 & 62.93 & 61.52 & 61.50 & 60.89 & \underline{58.63} & \textbf{53.96}\\
                & {NIQE$\downarrow$}    & 15.00 & 12.01 & 10.17 & 10.20 & 10.21 & 9.440 & 10.00 & 9.902 & \underline{9.011} & \textbf{8.371}\\
                & {PIQE$\downarrow$}    & 65.92 & 52.74 & 45.03 & 44.83 & 44.84 & 46.76 & 43.95 & \underline{43.51} & 44.63 & \textbf{41.53}\\
                & {RankIQA$\downarrow$} & 9.428 & 7.542 & 6.439 & 6.411 & 6.411 & 5.573 & 6.285 & 6.223 & \underline{5.320} & \textbf{4.895}\\
                & {MetaIQA$\uparrow$}  & .1241 & .1418 & .1732 & .1737 & .1737 & .1809 & .1773 & .1790 & \underline{.1909} & \textbf{.1969}\\
            \bottomrule
        \end{tabular}
    }
\end{table*}

The total loss function is the combination of the above losses, defined as:
\begin{equation}
\label{eq:total_loss}
\begin{aligned}
\mathcal{L}_{total} = & \lambda_{blur} \mathcal{L}_{blur} + \lambda_{event} \mathcal{L}_{event} \\
& + \lambda_{int} \mathcal{L}_{int} + \lambda_{depth} \mathcal{L}_{depth}.
\end{aligned}
\end{equation}

\section{Experiments} \label{sec:experiments}
\subsection{Implementation Details}
\label{subsec:implementation}
\textbf{Progressive Training}. 
We implemented EvaGaussians based on the official code of 3D-GS~\cite{kerbl3Dgaussians}. The training process spans 50,000 iterations, with an event reconstruction loss introduced after a 3,000-iteration warmup and we omit the densification process to streamline and simplify the subsequent optimization. 
Additionally, we adopt a coarse-to-fine training strategy, starting with rendering at a low resolution (0.3$\times$ downsampling in the early 30\% iterations) and progressively increasing the size of the rendered views to full resolution. 
All experiments were conducted using a single NVIDIA RTX 4090 GPU.

\noindent\textbf{Hyperparameter Setting}. 
During the training process, we set $\lambda_{1}=0.2$, $\lambda_{blur}=1.0$, $\lambda_{depth}=1.0e^{-2}$, $\lambda_{event}=5.0e^{-3}$ and $\lambda_{int}=1.0e^{-3}$ for the loss function, and used $n=9$ for the number of poses to be optimized during the exposure time. 
In implementing the loss $\mathcal{L}_{event}$, we configured the positive threshold as ${c}_{pos} = 0.25$ and the negative threshold as ${c}_{neg} = 0.25$ for synthetic scenes, and set ${c}_{pos} = 0.197$ and ${c}_{neg} = 0.241$ for real scenes.

\subsection{Datasets}
To facilitate a comprehensive evaluation, we introduce two novel datasets, with an overview provided below. Detailed information is presented in the supplementary.

\noindent\textbf{EvaGaussians-Blender Dataset.}
We construct a synthetic dataset covering a variety of scene scales, coupling with diverse camera trajectories and event data.
For large-scale scenes, we employ Blender to craft five distinct scenes, including city blocks and natural landscapes.
For medium-scale scenes, we craft three scenes using Blender, and redesign the camera trajectories of four scenes from DeblurNeRF~\cite{deblurnerf}. 
For object-level scenes, we create six scenes based on the NeRF-synthetic~\cite{nerf} dataset. 
We simulate motion blur by manually placing multi-view cameras, randomly adjusting camera poses, and performing linear interpolation between the original and perturbed positions for each view. The images are rendered from these interpolated poses and blended in RGB space to produce the final blurry images. 
The corresponding event streams are simulated using ESIM~\cite{esim} and V2E~\cite{vid2e}. 
The resulting large-scale and medium-scale scenes comprise 35 views of blurry images along with their corresponding event data, whereas the object-level scenes feature 100 views of blurry images.

\begin{figure*}[t]
    \begin{center}
    \includegraphics[width=1\linewidth,trim={0.0cm 0.0cm 0.0cm 0.2cm},clip]{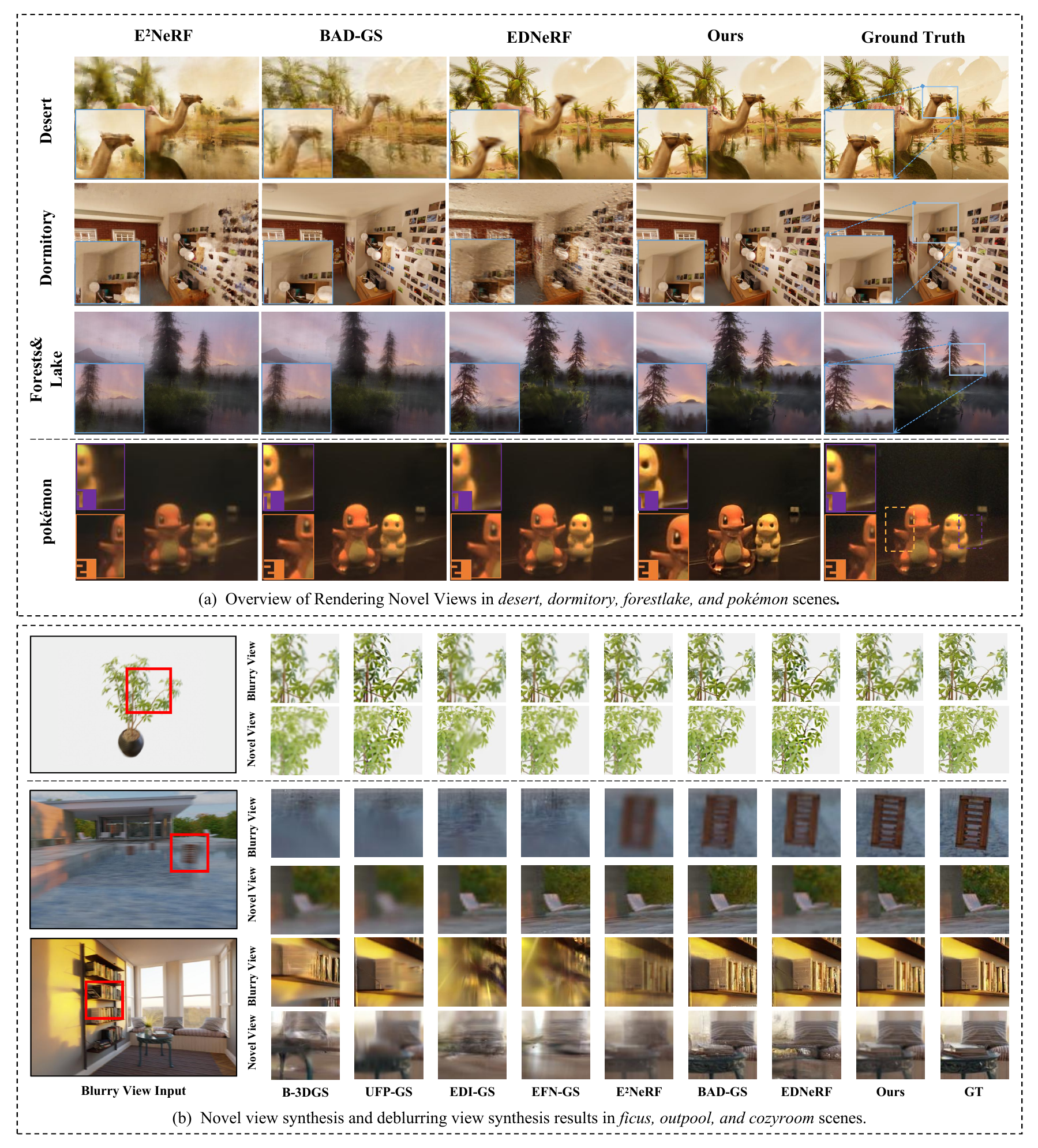}  
    \end{center}
    \vspace{-1.2em}
    \caption{Qualitative comparison on the synthetic and real dataset. We show the rendering novel views on the top section (a) and exhibit both novel view synthesis results and input view deblurring results on the bottom section (b). It shows that our method achieves better performance in recovering the training blurry views as well as rendering novel views. More results are presented in the supplementary.}
    \vspace{-0.65em}
    \label{fig:comparison—synthetic-scenes}
\end{figure*}

\noindent\textbf{EvaGaussians-DAVIS Dataset.} 
We manually recorded five real-world scenes using the Color DAVIS346 event camera~\cite{davis}, which has a resolution of $346\times260$ pixels and an exposure time of 100 milliseconds for the RGB frames. The dataset includes three object-level scenes and two indoor scenes. After processing, the final dataset consists of 30 images per scene, along with the recorded event streams, each showcasing various blur and lighting conditions.

\subsection{Experiment Settings}
\textbf{Baselines.} 
We compare our method with three types of baselines: 
1)~NeRF~\cite{nerf} and 3D-GS~\cite{kerbl3Dgaussians} directly trained on the blurry images, referring to as B-NeRF and B-3DGS. 
2)~Deblur rendering methods, including BAD-NeRF~\cite{wang2023badnerf}, BAD-GS~\cite{zhao2024badgs}, \({\textnormal{E}^2}\)\textnormal{NeRF}~\cite{qi2023e2nerf}, and EDNeRF~\cite{evdeblurnerf}. Among these, the first two methods simulate motion blur and optimize camera trajectories without event stream, whereas the latter two are event-assisted methods without camera trajectory optimization.
3)~Image deblur methods, including UFP~\cite{ufp} (single-image deblurring), EDI~\cite{pan2019edi} (event-based deblurring), and EFNet~\cite{efnet} (learnable event-based deblurring). We process input blurry images with them and train the vanilla 3D-GS with pre-deblurred images. The resulting baselines are referred to as UFP-GS, EDI-GS, and EFN-GS.

\noindent\textbf{Evaluation Metrics.} 
For synthetic datasets, we employ the Peak Signal-to-Noise Ratio (PSNR), Structural Similarity Index Measure (SSIM)~\cite{wang2004image}, and VGG-based Learned Perceptual Image Patch Similarity (LPIPS)~\cite{zhang2018unreasonable} to evaluate the similarity between rendered novel views and ground-truth novel views. 
For real-world datasets, since the sharp ground-truth images are unavailable, we utilize several No-Reference Image Quality Assessment (NR-IQA) metrics for evaluation, including BRISQUE~\cite{mittal2012no}, NIQE~\cite{mittal2012making}, PIQE~\cite{venkatanath2015blind}, RankIQA~\cite{liu2017rankiqa}, and MetaIQA~\cite{zhu2020Metaiqa}, which allow for image evaluation when lacking ground truth images. 

\subsection{Synthetic Data Experiments}
\label{sec4.3:Synthetic Data Experiments}
We evaluate our approach across a variety of scenes, including large-scale scenes, medium-scale scenes, and object-level scenes. Quantitative assessments of novel view synthesis are shown in the first three rows of
Table.~\ref{tab:quan_table}. The deblurring results of input views are detailed in the supplementary.
It can be found that our method achieves substantial improvements in most of the metrics,
especially in challenging large scenes. 
Specifically, both B-NeRF and B-3DGS produce blurry novel views since they are directly trained on blurred images. The image deblurring-based baselines, UFP-GS, EDI-GS and EFN-GS, also produced inferior results, because the image deblurring process potentially corrupts the 3D consistency of the training images.
Notably, our approach outperforms BAD-GS~\cite{zhao2024badgs} and BAD-NeRF~\cite{wang2023badnerf}, due to their limited capability in modeling complex textures. 
In addition, our method also surpasses the event-assisted methods \({\textnormal{E}^2}\)\textnormal{NeRF}~\cite{qi2023e2nerf} and EDNeRF~\cite{evdeblurnerf} in producing high-quality novel views with intricate details, with better training and rendering efficiency. An extended analysis of all the baselines is provided in the supplementary.

The qualitative results are illustrated in Figure.~\ref{fig:comparison—synthetic-scenes}, where the first three rows of Figure.~\ref{fig:comparison—synthetic-scenes}(a) shows novel view synthesis results, and Figure.~\ref{fig:comparison—synthetic-scenes}(b) shows both novel view and deblurring view synthesis results. More visualization results are provided in the supplementary.
It can be found that although
\({\textnormal{E}^2}\)\textnormal{NeRF}~\cite{qi2023e2nerf} performs well in object-level scenes, it struggles in medium and large-scale scene modeling, producing significant blurring results. 
Additionally, BAD-GS~\cite{zhao2024badgs} falls short in regions with significant color and depth variations, and produces overly smooth background textures.
Although EDNeRF~\cite{evdeblurnerf} exhibits overall satisfactory performance, its complex network architecture prolongs the training time (about 7 hours per scene) and precludes real-time rendering. In comparison, our method overcomes the baselines in producing high-fidelity novel views, and significantly reducing training time as well as demonstrating substantial advantages in real-time application scenarios. 


\begin{table*}[!ht]
\centering
\caption{Quantitative ablation on proposed loss functions. Best-performing results are highlighted in \textbf{bold} and second results in \underline{underline}.}
\label{tab:ablation_losses}
\resizebox{1.0\textwidth}{!}{ 
\begin{tabular}{cccc|ccc|ccc|ccccc}
\toprule
& & & & 
\multicolumn{3}{c|}{\textbf{Large-scale}} & 
\multicolumn{3}{c|}{\textbf{Medium-scale}} & 
\multicolumn{5}{c}{\textbf{Real Scene}} \\
\cmidrule(r){5-7} \cmidrule(r){8-10} \cmidrule(r){11-15}
 $\mathcal{L}_{\text{blur}}$ &  $\mathcal{L}_{\text{event}}$ &  $\mathcal{L}_{\text{depth}}$ & $\mathcal{L}_{\text{int}}$ & PSNR$\uparrow$ & SSIM$\uparrow$ & LPIPS$\downarrow$ & PSNR$\uparrow$ & SSIM$\uparrow$ & LPIPS$\downarrow$ & BRISQUE$\downarrow$ & NIQE$\downarrow$ & PIQE$\downarrow$ & MetaIQA$\uparrow$  & RankIQA$\downarrow$ \\
\midrule
\ding{51}& \ding{55} & \ding{55} & \ding{55} & 24.98 & .7865 &  .2986 & 29.10 & .8954 & .1673 & 57.32 & 8.799 & 43.26 & .1977 & 6.125\\
\ding{51}& \ding{51} & \ding{55} & \ding{55} & \underline{25.71} & .7949 &  .2745 & 29.94 & .9098 & .1551 & \underline{56.16} & 8.582 & 42.51 & .2009 &  5.066 \\
\ding{51}& \ding{51} & \ding{51} & \ding{55} & 25.54 & \underline{.8003} &  \underline{.2711} & \underline{30.05} & \underline{.9139} & \underline{.1542} & 56.17 & \underline{8.499} & \underline{42.23} & \underline{.1996} & \underline{4.923} \\
\ding{51}& \ding{51} & \ding{51} & \ding{51} & \textbf{26.02} & \textbf{.8064} &  \textbf{.2680} & \textbf{30.47} & \textbf{.9164} & \textbf{.1519} & \textbf{53.95} & \textbf{8.371} & \textbf{41.52} & \textbf{.2129} & \textbf{4.895} \\
\bottomrule
\end{tabular}
}
\end{table*}
\begin{table}[h]
\centering
 \setlength{\tabcolsep}{11pt}
\caption{Ablation study about the impact of pose optimization.}
\label{tab:ate_comparison}
\resizebox{1.\columnwidth}{!}{ 
\begin{tabular}{c|c|c}
\toprule
ATE $\downarrow$ & \textbf{Medium-scale} & \textbf{Large-scale}  \\
\midrule
\textbf{Initial poses} & $0.0598 \pm 0.0167$ & $0.3825 \pm 0.1245$  \\
\textbf{Optimized poses}  & $0.0461 \pm 0.0129$ & $0.0489 \pm 0.0112$  \\
\bottomrule
\end{tabular}
}
\end{table}
\begin{table}[t]
 \setlength{\tabcolsep}{12pt}
\centering
\caption{Robustness against motion blur level. 
}
\resizebox{0.48\textwidth}{!}{
\begin{tabular}{c|c|c|c}
\toprule
 Blur level & \textbf{Mild Blur} & \textbf{Medium Blur} & \textbf{Strong Blur} \\ \midrule
PSNR$ \uparrow$  & 26.38 & 25.71 & 25.04 \\ 
SSIM$ \uparrow$  & .8163 & .7949 & .7886 \\ 
LPIPS$ \downarrow$ & .2694 & .2745 & .2802 \\ 
\bottomrule
\end{tabular}
}
\label{tab:motion_blur}
\end{table}
\subsection{Real-world Data Experiments}
We present the quantitative results on the captured real-world data in the last row of Table.~\ref{tab:quan_table}. 
It can be found that our method achieves superior performance compared to other approaches. Specifically, for NR-IQA metrics, we achieve improvements in BRISQUE~\cite{mittal2012no}, NIQE~\cite{mittal2012making}, PIQE~\cite{venkatanath2015blind}, and RankIQA~\cite{liu2017rankiqa} by 15.38\%, 19.50\%, 11.49\%, and 22.83\% respectively. We also achieve an increase in 19.38\% in MetaIQA~\cite{zhu2020Metaiqa}.
The qualitative comparisons are shown in the last row of Figure.~\ref{fig:comparison—synthetic-scenes}(a) and in the supplementary, which further demonstrate that our method is capable of reconstructing detailed textures, ultimately achieving higher-quality novel view synthesis.

\subsection{Ablation Study}

\begin{figure}[t]
\centering
\includegraphics[width=.98\columnwidth]{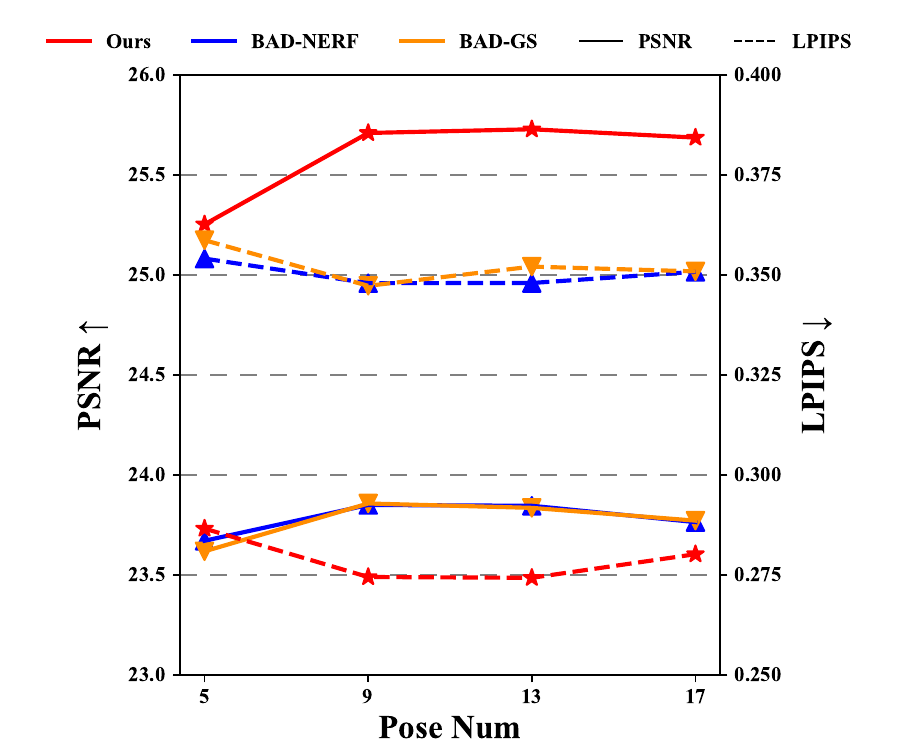}
\caption{Ablation on number of poses in the camera trajectory. }
\label{fig:ablatio_pose}
\end{figure}

\textbf{Camera Poses Optimization.}  
We firstly conduct ablations to investigate the effect of the number of camera poses optimized in the exposure time. We select five large scenes from our synthetic dataset for evaluation. In the experiments, we vary the number of camera poses, denoted as \(n\), from 5, 9, 13, and 17. The quantitative results of the novel view rendering are displayed in Figure.~\ref{fig:ablatio_pose}. It indicates that the results reach a bottleneck at 9 poses. Beyond this point, the improvements are limited and may potentially lead to local convergence issues. Based on these experiments, we choose \(n = 9\) camera poses to achieve a balance between rendering performance and training efficiency. Here, we also provide comparison with BAD-NeRF~\cite{wang2023badnerf} and BAD-GS~\cite{zhao2024badgs}. These two methods typically use linear interpolation to obtain camera trajectory, while our camera trajectories are estimated from the decomposed latent images, which provides more accurate initialization and helps our method achieves better performance. 
Moreover, we conduct quantitative experiments using 9 camera poses to compute ATE (Average Trajectory Error) of the initial poses produced by COLMAP~\cite{colmap} and the optimized poses, the results are shown in Table.~\ref{tab:ate_comparison}, which validates the effectiveness of pose optimization. 

\noindent\textbf{Effectiveness of The Loss Functions.}  
We conduct novel view synthesis experiments on the proposed datasets to validate the effectiveness of the training losses. The quantitative results, as shown in Table.~\ref{tab:ablation_losses}, indicate that using only the blur reconstruction loss leads to suboptimal outputs, performing poorly and lacking high-frequency details  on both synthetic and real-world datasets.
In contrast, incorporating $\mathcal{L}_{\text{event}}$, $\mathcal{L}_{\text{depth}}$, and $\mathcal{L}_{\text{int}}$ enables our proposed method to produce high-fidelity novel views with intricate details.

\noindent\textbf{Robustness Against Motion Blur Levels.}  
To validate the robustness of our method in handling different levels of motion blur, we set up three different camera speeds in the \textit{city blocks scene} of the synthetic dataset to obtain images with varying degrees of blur. Images with mild blur are captured at half the default camera motion speed, images with medium blur are captured at the default motion speed, and images with strong blur are captured at twice the default motion speed. The quantitative results of novel view synthesis are listed in Table.~\ref{tab:motion_blur}. It can be observed that the results show no significant fluctuations across different levels of blur, demonstrating the robustness of our method to varying motion blur levels. Please refer to the supplementary for more ablations of our method.

\section{Conclusions} \label{sec:conclusions}
This paper introduces Event Stream Assisted Gaussian Splatting (\textbf{EvaGaussians}), a novel framework that seamlessly integrates the event streams captured by an event camera into the training of 3D-GS, effectively addressing the challenges of reconstructing high-quality 3D-GS from motion-blurred images. We contribute two novel datasets and conduct comprehensive experiments. The results demonstrate that our method outperforms previous state-of-the-art deblurring rendering techniques in terms of novel view synthesis quality, without sacrificing inference efficiency.
Despite its promising performance, our method may still face challenges when reconstructing scenes with extremely intricate textures from severely blurred images. We will release our code and dataset for future research.


\clearpage
{
    \small
    \bibliographystyle{ieeenat_fullname}
    \bibliography{main}

\begin{thebibliography}{54}
\providecommand{\natexlab}[1]{#1}
\providecommand{\url}[1]{\texttt{#1}}
\expandafter\ifx\csname urlstyle\endcsname\relax
  \providecommand{\doi}[1]{doi: #1}\else
  \providecommand{\doi}{doi: \begingroup \urlstyle{rm}\Url}\fi

\bibitem[Brandli et~al.(2014)Brandli, Muller, and Delbruck]{davis346}
Christian Brandli, Lorenz Muller, and Tobi Delbruck.
\newblock Real-time, high-speed video decompression using a frame- and event-based davis sensor.
\newblock In \emph{2014 IEEE International Symposium on Circuits and Systems (ISCAS)}, pages 686--689, 2014.

\bibitem[Cannici and Scaramuzza(2024{\natexlab{a}})]{cannici2024mitigating}
Marco Cannici and Davide Scaramuzza.
\newblock Mitigating motion blur in neural radiance fields with events and frames.
\newblock In \emph{CVPR}, 2024{\natexlab{a}}.

\bibitem[Cannici and Scaramuzza(2024{\natexlab{b}})]{evdeblurnerf}
Marco Cannici and Davide Scaramuzza.
\newblock Mitigating motion blur in neural radiance fields with events and frames.
\newblock In \emph{Proceedings of the IEEE/CVF Conference on Computer Vision and Pattern Recognition (CVPR)}, 2024{\natexlab{b}}.

\bibitem[Comi et~al.(2024)Comi, Tonioni, Yang, Tremblay, Blukis, Lin, Lepora, and Aitchison]{comi2024snap}
Mauro Comi, Alessio Tonioni, Max Yang, Jonathan Tremblay, Valts Blukis, Yijiong Lin, Nathan~F Lepora, and Laurence Aitchison.
\newblock Snap-it, tap-it, splat-it: Tactile-informed 3d gaussian splatting for reconstructing challenging surfaces.
\newblock \emph{arXiv preprint arXiv:2403.20275}, 2024.

\bibitem[Community(2018)]{Blender}
Blender~Online Community.
\newblock \emph{{Blender - a 3D modelling and rendering package}}.
\newblock Blender Foundation, Stichting Blender Foundation, Amsterdam, 2018.

\bibitem[Fang et~al.(2023)Fang, Wu, Dong, Li, Wu, and Shi]{ufp}
Zhenxuan Fang, Fangfang Wu, Weisheng Dong, Xin Li, Jinjian Wu, and Guangming Shi.
\newblock Self-supervised non-uniform kernel estimation with flow-based motion prior for blind image deblurring.
\newblock In \emph{Proceedings of the IEEE/CVF Conference on Computer Vision and Pattern Recognition}, pages 18105--18114, 2023.

\bibitem[Gallego et~al.(2020)Gallego, Delbr{\"u}ck, Orchard, Bartolozzi, Taba, Censi, Leutenegger, Davison, Conradt, Daniilidis, et~al.]{gallego2020survey}
Guillermo Gallego, Tobi Delbr{\"u}ck, Garrick Orchard, Chiara Bartolozzi, Brian Taba, Andrea Censi, Stefan Leutenegger, Andrew~J Davison, J{\"o}rg Conradt, Kostas Daniilidis, et~al.
\newblock Event-based vision: A survey.
\newblock \emph{IEEE TPAMI}, 2020.

\bibitem[Heise et~al.(2013)Heise, Klose, Jensen, and Knoll]{heise2013pm}
Philipp Heise, Sebastian Klose, Brian Jensen, and Alois Knoll.
\newblock Pm-huber: Patchmatch with huber regularization for stereo matching.
\newblock In \emph{ICCV}, 2013.

\bibitem[Hu et~al.(2021)Hu, Liu, and Delbruck]{vid2e}
Yuhuang Hu, Shih-Chii Liu, and Tobi Delbruck.
\newblock v2e: From video frames to realistic dvs events.
\newblock In \emph{Proceedings of the IEEE/CVF Conference on Computer Vision and Pattern Recognition}, pages 1312--1321, 2021.

\bibitem[Hwang et~al.(2023)Hwang, Kim, and Kim]{hwang2023ev}
Inwoo Hwang, Junho Kim, and Young~Min Kim.
\newblock Ev-nerf: Event based neural radiance field.
\newblock In \emph{Proceedings of the IEEE/CVF Winter Conference on Applications of Computer Vision}, pages 837--847, 2023.

\bibitem[Jiang et~al.(2020)Jiang, Zhang, Zou, Ren, Lv, and Liu]{jiang2020emd}
Zhe Jiang, Yu Zhang, Dongqing Zou, Jimmy Ren, Jiancheng Lv, and Yebin Liu.
\newblock Learning event-based motion deblurring.
\newblock In \emph{CVPR}, 2020.

\bibitem[Jin et~al.(2024)Jin, Zhu, Yuan, and Yan]{jinpeng2024moh}
Peng Jin, Bo Zhu, Li Yuan, and Shuicheng Yan.
\newblock Moh: Multi-head attention as mixture-of-head attention.
\newblock \emph{arXiv preprint arXiv:2410.11842}, 2024.

\bibitem[Kerbl et~al.(2023)Kerbl, Kopanas, Leimk{\"u}hler, and Drettakis]{kerbl3Dgaussians}
Bernhard Kerbl, Georgios Kopanas, Thomas Leimk{\"u}hler, and George Drettakis.
\newblock 3d gaussian splatting for real-time radiance field rendering.
\newblock \emph{ACM TOG}, 2023.

\bibitem[Kerr et~al.(2023)Kerr, Kim, Goldberg, Kanazawa, and Tancik]{kerr2023lerf}
Justin Kerr, Chung~Min Kim, Ken Goldberg, Angjoo Kanazawa, and Matthew Tancik.
\newblock Lerf: Language embedded radiance fields.
\newblock In \emph{CVPR}, 2023.

\bibitem[Klenk et~al.(2023)Klenk, Koestler, Scaramuzza, and Cremers]{enerf}
Simon Klenk, Lukas Koestler, Davide Scaramuzza, and Daniel Cremers.
\newblock E-nerf: Neural radiance fields from a moving event camera.
\newblock \emph{IEEE Robotics and Automation Letters}, 2023.

\bibitem[LaViola~Jr(2008)]{laviola2008bringing}
Joseph~J LaViola~Jr.
\newblock Bringing vr and spatial 3d interaction to the masses through video games.
\newblock \emph{IEEE Computer Graphics and Applications}, 2008.

\bibitem[Lee et~al.(2023)Lee, Lee, Shin, and Lee]{lee2023dp}
Dogyoon Lee, Minhyeok Lee, Chajin Shin, and Sangyoun Lee.
\newblock Dp-nerf: Deblurred neural radiance field with physical scene priors.
\newblock In \emph{CVPR}, 2023.

\bibitem[Lin et~al.(2020)Lin, Zhang, Pan, Jiang, Zou, Wang, Chen, and Ren]{lin2020edv}
Songnan Lin, Jiawei Zhang, Jinshan Pan, Zhe Jiang, Dongqing Zou, Yongtian Wang, Jing Chen, and Jimmy Ren.
\newblock Learning event-driven video deblurring and interpolation.
\newblock In \emph{ECCV}, 2020.

\bibitem[Liu et~al.(2023)Liu, Zhan, Zhang, Xu, Yu, El~Saddik, Theobalt, Xing, and Lu]{liu2023weakly}
Kunhao Liu, Fangneng Zhan, Jiahui Zhang, Muyu Xu, Yingchen Yu, Abdulmotaleb El~Saddik, Christian Theobalt, Eric Xing, and Shijian Lu.
\newblock Weakly supervised 3d open-vocabulary segmentation.
\newblock In \emph{NeurIPS}, 2023.

\bibitem[Liu et~al.(2024)Liu, Shi, Kuang, Zhu, Li, Han, Cai, Porikli, and Su]{liu2024openshape}
Minghua Liu, Ruoxi Shi, Kaiming Kuang, Yinhao Zhu, Xuanlin Li, Shizhong Han, Hong Cai, Fatih Porikli, and Hao Su.
\newblock Openshape: Scaling up 3d shape representation towards open-world understanding.
\newblock In \emph{NeurIPS}, 2024.

\bibitem[Liu et~al.(2017)Liu, Van De~Weijer, and Bagdanov]{liu2017rankiqa}
Xialei Liu, Joost Van De~Weijer, and Andrew~D Bagdanov.
\newblock Rankiqa: Learning from rankings for no-reference image quality assessment.
\newblock In \emph{Proceedings of the IEEE international conference on computer vision}, pages 1040--1049, 2017.

\bibitem[Low and Lee(2023)]{low2023robust}
Weng~Fei Low and Gim~Hee Lee.
\newblock Robust e-nerf: Nerf from sparse \& noisy events under non-uniform motion.
\newblock In \emph{Proceedings of the IEEE/CVF International Conference on Computer Vision}, 2023.

\bibitem[Low and Lee(2025)]{low2025deblur-e-nerf}
Weng~Fei Low and Gim~Hee Lee.
\newblock Deblur e-nerf: Nerf from motion-blurred events under high-speed or low-light conditions.
\newblock In \emph{European Conference on Computer Vision}, pages 192--209. Springer, 2025.

\bibitem[Ma et~al.(2022{\natexlab{a}})Ma, Li, Liao, Zhang, Wang, Wang, and Sander]{deblur-nerf}
Li Ma, Xiaoyu Li, Jing Liao, Qi Zhang, Xuan Wang, Jue Wang, and Pedro~V Sander.
\newblock {Deblur-NeRF: Neural Radiance Fields from Blurry Images}.
\newblock In \emph{CVPR}, 2022{\natexlab{a}}.

\bibitem[Ma et~al.(2022{\natexlab{b}})Ma, Li, Liao, Zhang, Wang, Wang, and Sander]{deblurnerf}
Li Ma, Xiaoyu Li, Jing Liao, Qi Zhang, Xuan Wang, Jue Wang, and Pedro~V Sander.
\newblock Deblur-nerf: Neural radiance fields from blurry images.
\newblock In \emph{Proceedings of the IEEE/CVF Conference on Computer Vision and Pattern Recognition}, pages 12861--12870, 2022{\natexlab{b}}.

\bibitem[Mildenhall et~al.(2020)Mildenhall, Srinivasan, Tancik, Barron, Ramamoorthi, and Ng]{nerf}
Ben Mildenhall, Pratul~P Srinivasan, Matthew Tancik, Jonathan~T Barron, Ravi Ramamoorthi, and Ren Ng.
\newblock {NeRF: Representing Scenes as Neural Radiance Fields for View Synthesis}.
\newblock In \emph{ECCV}, 2020.

\bibitem[Mittal et~al.(2012{\natexlab{a}})Mittal, Moorthy, and Bovik]{mittal2012no}
Anish Mittal, Anush~Krishna Moorthy, and Alan~Conrad Bovik.
\newblock No-reference image quality assessment in the spatial domain.
\newblock \emph{IEEE TIP}, 21\penalty0 (12):\penalty0 4695--4708, 2012{\natexlab{a}}.

\bibitem[Mittal et~al.(2012{\natexlab{b}})Mittal, Soundararajan, and Bovik]{mittal2012making}
Anish Mittal, Rajiv Soundararajan, and Alan~C Bovik.
\newblock Making a “completely blind” image quality analyzer.
\newblock \emph{IEEE Signal processing letters}, 20\penalty0 (3):\penalty0 209--212, 2012{\natexlab{b}}.

\bibitem[Pan et~al.(2019)Pan, Scheerlinck, Yu, Hartley, Liu, and Dai]{pan2019edi}
Liyuan Pan, Cedric Scheerlinck, Xin Yu, Richard Hartley, Miaomiao Liu, and Yuchao Dai.
\newblock Bringing a blurry frame alive at high frame-rate with an event camera.
\newblock In \emph{CVPR}, 2019.

\bibitem[Qi et~al.(2023)Qi, Zhu, Zhang, and Li]{qi2023e2nerf}
Yunshan Qi, Lin Zhu, Yu Zhang, and Jia Li.
\newblock E2nerf: Event enhanced neural radiance fields from blurry images.
\newblock In \emph{ICCV}, 2023.

\bibitem[Rebecq et~al.(2018)Rebecq, Gehrig, and Scaramuzza]{esim}
Henri Rebecq, Daniel Gehrig, and Davide Scaramuzza.
\newblock Esim: an open event camera simulator.
\newblock In \emph{Conference on robot learning}, pages 969--982. PMLR, 2018.

\bibitem[Rosinol et~al.(2023)Rosinol, Leonard, and Carlone]{rosinol2023nerfslam}
Antoni Rosinol, John~J Leonard, and Luca Carlone.
\newblock Nerf-slam: Real-time dense monocular slam with neural radiance fields.
\newblock In \emph{IROS}, 2023.

\bibitem[Rudnev et~al.(2023)Rudnev, Elgharib, Theobalt, and Golyanik]{rudnev2023eventnerf}
Viktor Rudnev, Mohamed Elgharib, Christian Theobalt, and Vladislav Golyanik.
\newblock Eventnerf: Neural radiance fields from a single colour event camera.
\newblock In \emph{Computer Vision and Pattern Recognition (CVPR)}, 2023.

\bibitem[Schonberger and Frahm(2016)]{colmap}
Johannes~L Schonberger and Jan-Michael Frahm.
\newblock {Structure-from-motion Revisited}.
\newblock In \emph{CVPR}, 2016.

\bibitem[Sun et~al.(2022)Sun, Sakaridis, Liang, Jiang, Yang, Sun, Ye, Wang, and Gool]{efnet}
Lei Sun, Christos Sakaridis, Jingyun Liang, Qi Jiang, Kailun Yang, Peng Sun, Yaozu Ye, Kaiwei Wang, and Luc~Van Gool.
\newblock Event-based fusion for motion deblurring with cross-modal attention.
\newblock In \emph{European Conference on Computer Vision}, pages 412--428. Springer, 2022.

\bibitem[Tang et~al.(2024{\natexlab{a}})Tang, Zhang, Cheng, Yu, Feng, Pang, Lin, and Yuan]{tang2024cycle3d}
Zhenyu Tang, Junwu Zhang, Xinhua Cheng, Wangbo Yu, Chaoran Feng, Yatian Pang, Bin Lin, and Li Yuan.
\newblock Cycle3d: High-quality and consistent image-to-3d generation via generation-reconstruction cycle.
\newblock \emph{arXiv preprint arXiv:2407.19548}, 2024{\natexlab{a}}.

\bibitem[Tang et~al.(2024{\natexlab{b}})Tang, Zhang, Cheng, Yu, Feng, Pang, Lin, and Yuan]{tangzhenyu2024cycle3d}
Zhenyu Tang, Junwu Zhang, Xinhua Cheng, Wangbo Yu, Chaoran Feng, Yatian Pang, Bin Lin, and Li Yuan.
\newblock Cycle3d: High-quality and consistent image-to-3d generation via generation-reconstruction cycle.
\newblock \emph{arXiv preprint arXiv:2407.19548}, 2024{\natexlab{b}}.

\bibitem[Taverni(2020)]{davis}
Gemma Taverni.
\newblock \emph{Applications of Silicon Retinas: From Neuroscience to Computer Vision}.
\newblock PhD thesis, Universit{\"a}t Z{\"u}rich, 2020.

\bibitem[Vairalkar and Nimbhorkar(2012)]{vairalkar2012edge-sobel}
Manoj~K Vairalkar and SU Nimbhorkar.
\newblock Edge detection of images using sobel operator.
\newblock \emph{International Journal of Emerging Technology and Advanced Engineering}, 2\penalty0 (1):\penalty0 291--293, 2012.

\bibitem[Venkatanath et~al.(2015)Venkatanath, Praneeth, Bh, Channappayya, and Medasani]{venkatanath2015blind}
N Venkatanath, D Praneeth, Maruthi~Chandrasekhar Bh, Sumohana~S Channappayya, and Swarup~S Medasani.
\newblock Blind image quality evaluation using perception based features.
\newblock In \emph{2015 twenty first national conference on communications (NCC)}, pages 1--6. IEEE, 2015.

\bibitem[Wang et~al.(2023)Wang, Zhao, Ma, and Liu]{wang2023badnerf}
Peng Wang, Lingzhe Zhao, Ruijie Ma, and Peidong Liu.
\newblock {BAD-NeRF: Bundle Adjusted Deblur Neural Radiance Fields}.
\newblock In \emph{CVPR}, 2023.

\bibitem[Wang et~al.(2004)Wang, Bovik, Sheikh, and Simoncelli]{wang2004image}
Zhou Wang, Alan~C Bovik, Hamid~R Sheikh, and Eero~P Simoncelli.
\newblock Image quality assessment: from error visibility to structural similarity.
\newblock \emph{IEEE transactions on image processing}, 13\penalty0 (4):\penalty0 600--612, 2004.

\bibitem[Wu et~al.(2024)Wu, Zhu, Wang, and Lam]{wu2024ev-gs}
Jingqian Wu, Shuo Zhu, Chutian Wang, and Edmund~Y Lam.
\newblock Ev-gs: Event-based gaussian splatting for efficient and accurate radiance field rendering.
\newblock In \emph{2024 IEEE 34th International Workshop on Machine Learning for Signal Processing (MLSP)}, pages 1--6. IEEE, 2024.

\bibitem[Xiong et~al.(2024)Xiong, Wu, He, Fermuller, Aloimonos, Huang, and Metzler]{xiong2024event3dgs}
Tianyi Xiong, Jiayi Wu, Botao He, Cornelia Fermuller, Yiannis Aloimonos, Heng Huang, and Christopher~A Metzler.
\newblock Event3dgs: Event-based 3d gaussian splatting for fast egomotion.
\newblock \emph{arXiv preprint arXiv:2406.02972}, 2024.

\bibitem[Xu et~al.(2023)Xu, Agrawal, Laney, Garcia, Bansal, Kim, Rota~Bul{\`o}, Porzi, Kontschieder, Bo{\v{z}}i{\v{c}}, et~al.]{xu2023vr}
Linning Xu, Vasu Agrawal, William Laney, Tony Garcia, Aayush Bansal, Changil Kim, Samuel Rota~Bul{\`o}, Lorenzo Porzi, Peter Kontschieder, Alja{\v{z}} Bo{\v{z}}i{\v{c}}, et~al.
\newblock Vr-nerf: High-fidelity virtualized walkable spaces.
\newblock In \emph{SIGGRAPH Asia}, 2023.

\bibitem[Yen-Chen et~al.(2021)Yen-Chen, Florence, Barron, Rodriguez, Isola, and Lin]{yen2021inerf}
Lin Yen-Chen, Pete Florence, Jonathan~T Barron, Alberto Rodriguez, Phillip Isola, and Tsung-Yi Lin.
\newblock inerf: Inverting neural radiance fields for pose estimation.
\newblock In \emph{IROS}, 2021.

\bibitem[Yin et~al.(2024)Yin, Shi, Bao, Bing, Liao, Yang, and Wang]{yin2024e-3dgs}
Xiaoting Yin, Hao Shi, Yuhan Bao, Zhenshan Bing, Yiyi Liao, Kailun Yang, and Kaiwei Wang.
\newblock E-3dgs: Gaussian splatting with exposure and motion events.
\newblock \emph{arXiv preprint arXiv:2410.16995}, 2024.

\bibitem[Yu et~al.(2024)Yu, Xing, Yuan, Hu, Li, Huang, Gao, Wong, Shan, and Tian]{yu2024viewcrafter}
Wangbo Yu, Jinbo Xing, Li Yuan, Wenbo Hu, Xiaoyu Li, Zhipeng Huang, Xiangjun Gao, Tien-Tsin Wong, Ying Shan, and Yonghong Tian.
\newblock Viewcrafter: Taming video diffusion models for high-fidelity novel view synthesis.
\newblock \emph{arXiv preprint arXiv:2409.02048}, 2024.

\bibitem[Yuan et~al.(2024)Yuan, Huang, Xu, Liu, Zhang, Shi, Zhu, Cheng, Luo, and Yuan]{yuanshenghai2024chronomagic}
Shenghai Yuan, Jinfa Huang, Yongqi Xu, Yaoyang Liu, Shaofeng Zhang, Yujun Shi, Ruijie Zhu, Xinhua Cheng, Jiebo Luo, and Li Yuan.
\newblock Chronomagic-bench: A benchmark for metamorphic evaluation of text-to-time-lapse video generation.
\newblock \emph{arXiv preprint arXiv:2406.18522}, 2024.

\bibitem[Zhang et~al.(2018)Zhang, Isola, Efros, Shechtman, and Wang]{zhang2018unreasonable}
Richard Zhang, Phillip Isola, Alexei~A Efros, Eli Shechtman, and Oliver Wang.
\newblock The unreasonable effectiveness of deep features as a perceptual metric.
\newblock In \emph{Proceedings of the IEEE conference on computer vision and pattern recognition}, pages 586--595, 2018.

\bibitem[Zhang et~al.(2024)Zhang, Chen, and Wang]{zhang2024elite-ev-gs}
Zixin Zhang, Kanghao Chen, and Lin Wang.
\newblock Elite-evgs: Learning event-based 3d gaussian splatting by distilling event-to-video priors.
\newblock \emph{arXiv preprint arXiv:2409.13392}, 2024.

\bibitem[Zhao et~al.(2024)Zhao, Wang, and Liu]{zhao2024badgs}
Lingzhe Zhao, Peng Wang, and Peidong Liu.
\newblock Bad-gaussians: Bundle adjusted deblur gaussian splatting.
\newblock In \emph{ECCV}, 2024.

\bibitem[Zhu et~al.(2020)Zhu, Li, Wu, Dong, and Shi]{zhu2020Metaiqa}
Hancheng Zhu, Leida Li, Jinjian Wu, Weisheng Dong, and Guangming Shi.
\newblock Metaiqa: Deep meta-learning for no-reference image quality assessment.
\newblock In \emph{Proceedings of the IEEE/CVF conference on computer vision and pattern recognition}, pages 14143--14152, 2020.

\bibitem[Zhu et~al.(2022)Zhu, Peng, Larsson, Xu, Bao, Cui, Oswald, and Pollefeys]{zhu2022nice}
Zihan Zhu, Songyou Peng, Viktor Larsson, Weiwei Xu, Hujun Bao, Zhaopeng Cui, Martin~R Oswald, and Marc Pollefeys.
\newblock Nice-slam: Neural implicit scalable encoding for slam.
\newblock In \emph{CVPR}, 2022.

\end{thebibliography}
}

\clearpage
\newcommand{\nocontentsline}[3]{}
\newcommand{\tocless}[2]{\bgroup\let\addcontentsline=\nocontentsline#1{#2}\egroup}

\newcommand{\Appendix}[1]{
  \refstepcounter{section}
  \section*{Appendix \thesection:\hspace*{1.5ex} #1}
  \addcontentsline{toc}{section}{Appendix \thesection}
}
\newcommand{\SubAppendix}[1]{\tocless\subsection{#1}}
\maketitlesupplementary
\appendix

\vspace{4mm}
\section{Dataset Details}
\label{app:dataset}

\subsection{EvaGaussians-Blender Dataset}
\subsubsection{Dataset Overview}

\begin{figure*}[h]
    \begin{center}
    \includegraphics[width=.95\linewidth]{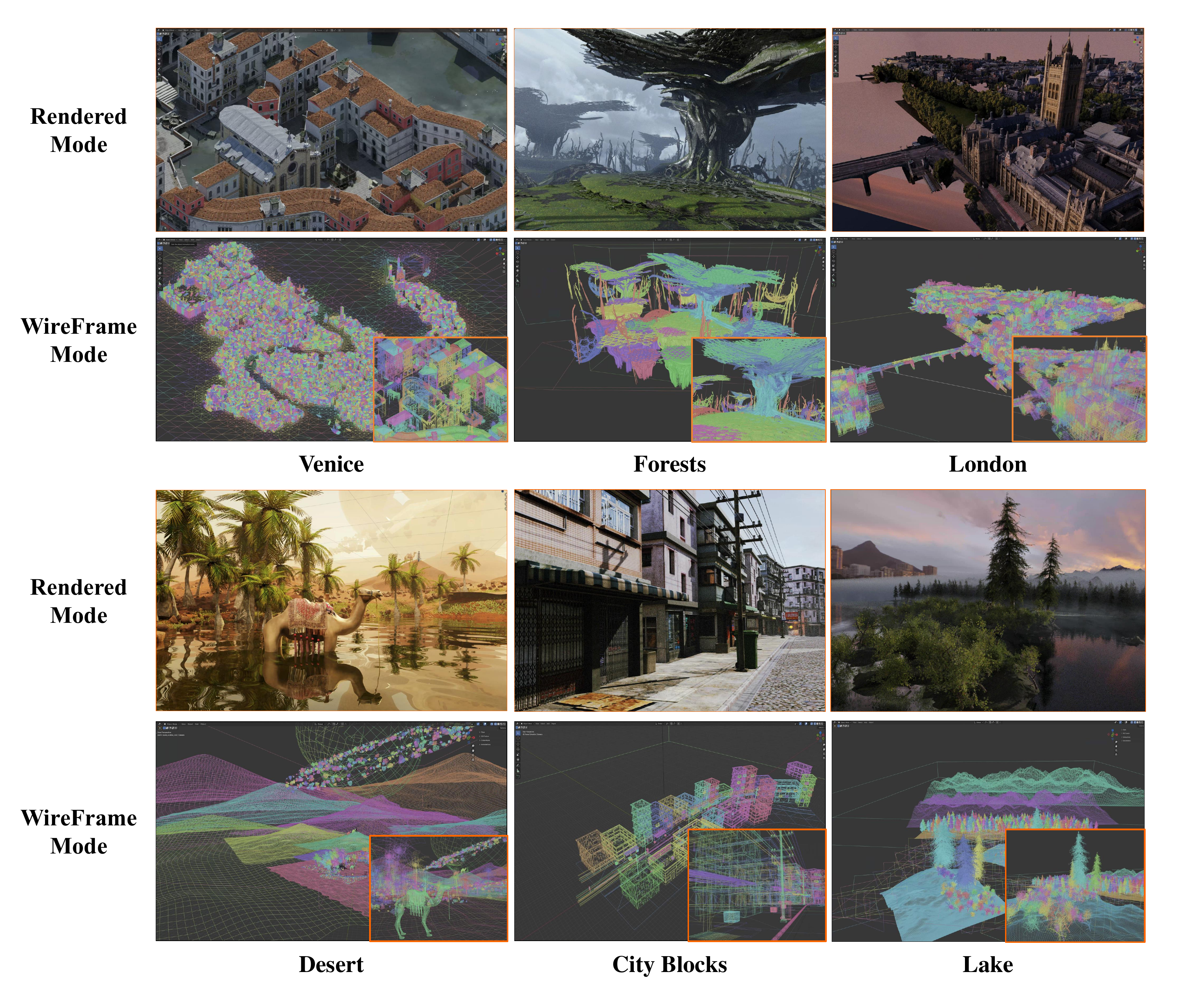}
    \end{center}
    \caption{Visualization of EvaGaussians-Blender Outdoor Scenes. These scenes include rich details ane diverse components like sky, lake, river, desert, forest, cities, roads. All scenes cover an area of more than 1 square kilometer.}
    \label{fig:append_outdoors}
\end{figure*}

We use Blender~\cite{Blender} to craft nine indoor and outdoor 3D scenes, and further incorporate four 3D scenes from DeblurNeRF~\cite{deblur-nerf} and six 3D objects from the NeRF-Synthetic dataset~\cite{nerf} as our base scenes. We then design various camera trajectories to simulate motion-blurred images on these base scenes, and generate the corresponding event streams using V2E~\cite{vid2e}. 
Visualization of our crafted scenes are shown in Figure.~\ref{fig:append_indoors} and Figure.~\ref{fig:append_outdoors}, and an overview of the design purposes of these crafted scenes is provided below:
\begin{figure*}[h]
    \begin{center}
    \includegraphics[width=.95\linewidth]{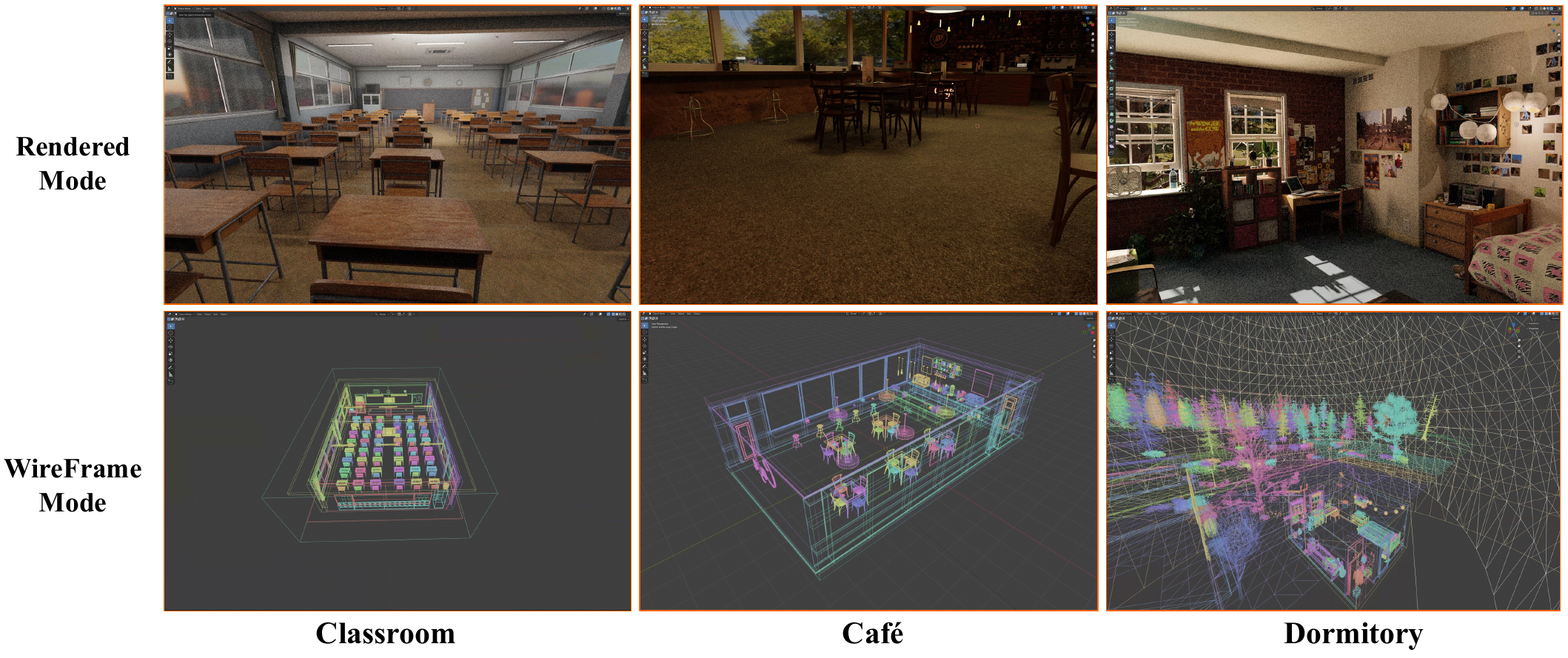}
    \end{center}
    \caption{Visualization of EvaGaussians-Blender Indoor Scenes. The sizes of the \textit{Café} and \textit{Classroom} scenes are approximately ${15\times7\times4}$ meters, while the \textit{Dormitory} scene is approximately ${5\times5\times4}$ meters (with an additional outdoor garden, making the overall scene size ${20\times20\times6}$ meters).}
    \label{fig:append_indoors}
\end{figure*}
\begin{figure}[h]
    \begin{center}
    \includegraphics[width=1.\linewidth]{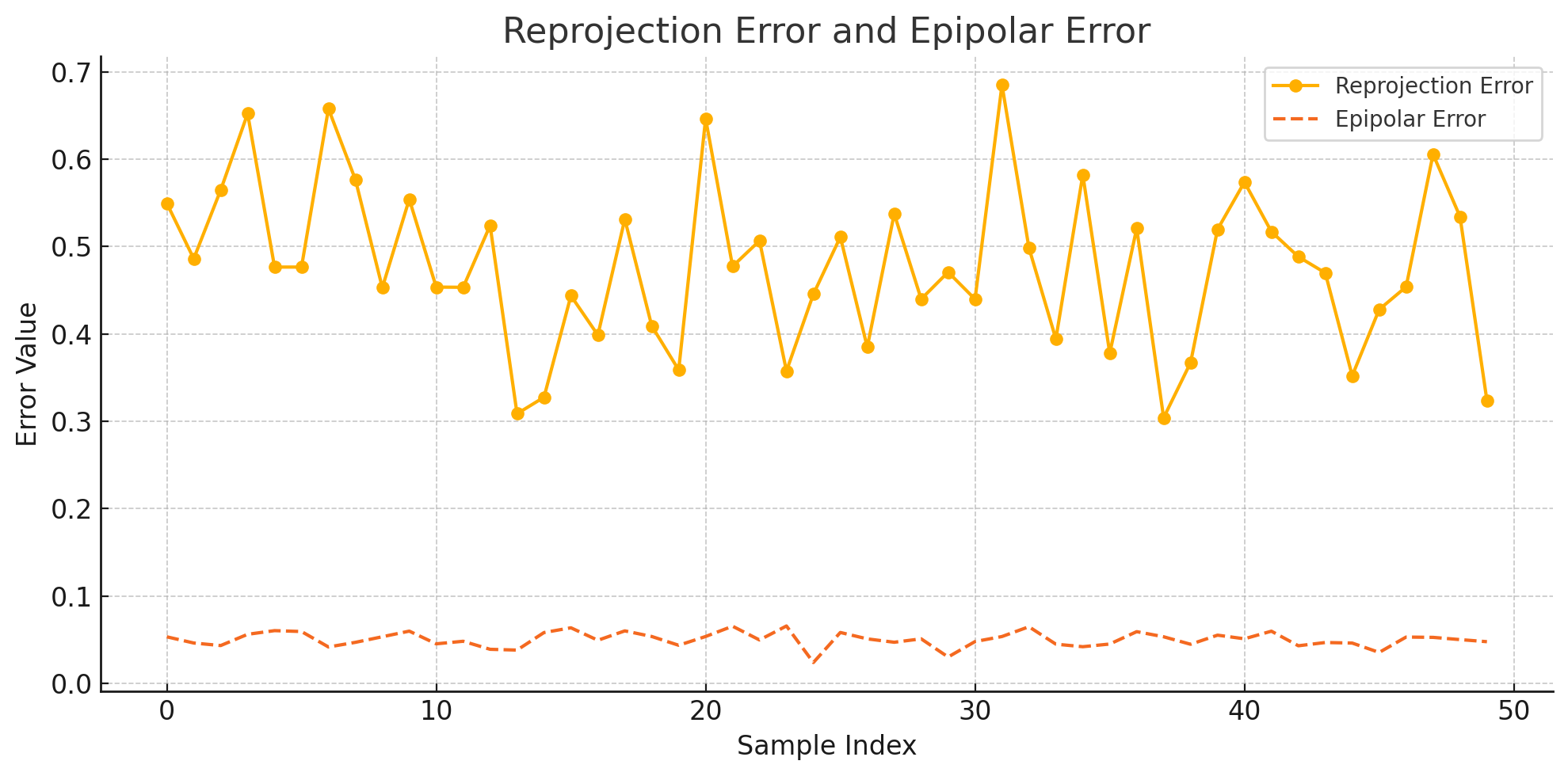}
    \end{center}
    \caption{Visualization of Reprojection Errors and Epipolar Errors. The figure illustrates the 50 sets of reprojection errors and epipolar errors generated during the calibration process. The reprojection error $ e_r $ represents the average discrepancy between the observed points and the projected points, calculated as shown in Eq.~\ref{eq:reprojection_error}. The epipolar error  $e_{\text{epipolar}}$ represents the average distance between points in one camera and the epipolar lines calculated from the other camera for each pair of images, calculated as shown in Eq.~\ref{eq:epipolar_error}. As shown in the figure, the average reprojection error is approximately 0.5, and the average epipolar error is approximately 0.05, indicating a high level of accuracy in the calibration process.}
    \label{fig:appendix_error}
\end{figure}
\begin{figure}
    \begin{center}
    \includegraphics[width=1.\linewidth]{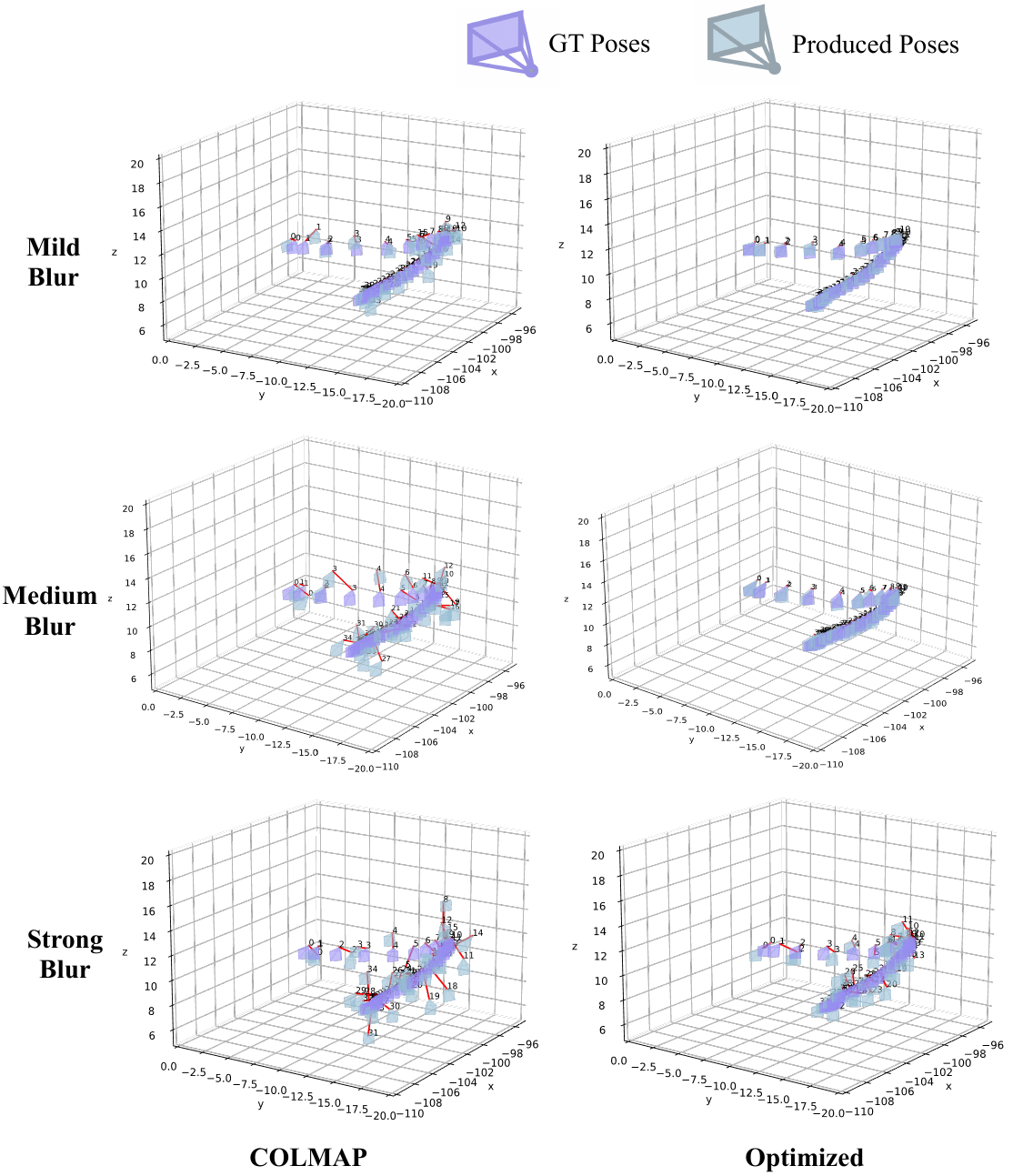}
    \end{center}
    \caption{Visualization of pose accuracy in different level of motion blur.}
    \label{fig:pose}
\end{figure}
\paragraph{Indoor Scenes}
\begin{itemize}
    \item \textit{Classroom}: A typical classroom setting featuring desks, chairs, a blackboard, and educational posters. This scene is designed to simulate an academic environment, ideal for educational and surveillance applications.
    \item \textit{Café}: A cozy café with tables, chairs, a counter, and various decorations. This scene mimics a social setting, providing a dynamic backdrop for testing social interaction algorithms and retail analytics.
    \item \textit{Dormitory}: A student dormitory room equipped with beds, study desks, personal belongings, and typical dorm furniture. This scene represents a personal living space, useful for smart home and security applications.
\end{itemize}

\paragraph{Outdoor Scenes}
\begin{itemize}
    \item \textit{Desert}: A vast, arid landscape with sand dunes and sparse vegetation. This scene is perfect for testing navigation and object detection in harsh, unstructured environments.
    \item \textit{City Blocks}: Urban scenes featuring streets, buildings, vehicles, and pedestrians. This environment is essential for autonomous driving, urban planning, and smart city applications.
    \item \textit{Lake}: A serene natural setting with dense forests surrounding a tranquil lake. This scene provides a complex environment for testing outdoor navigation, environmental monitoring, and wildlife tracking.
    \item \textit{Forests}: A rugged terrain with forested areas and scattered boulders. This scene is useful for off-road navigation and geological survey applications.
    \item \textit{Venice}: A picturesque representation of Venice with canals, bridges, and historic architecture. This scene offers a unique setting for cultural heritage preservation, tourism, and urban analytics.
    \item \textit{London}: A bustling cityscape of London with iconic landmarks, streets, and a dynamic urban environment. This scene supports applications in tourism, traffic management, and city modeling.
\end{itemize}

\begin{figure*}[h]
    \begin{center}
    \includegraphics[width=1\linewidth]{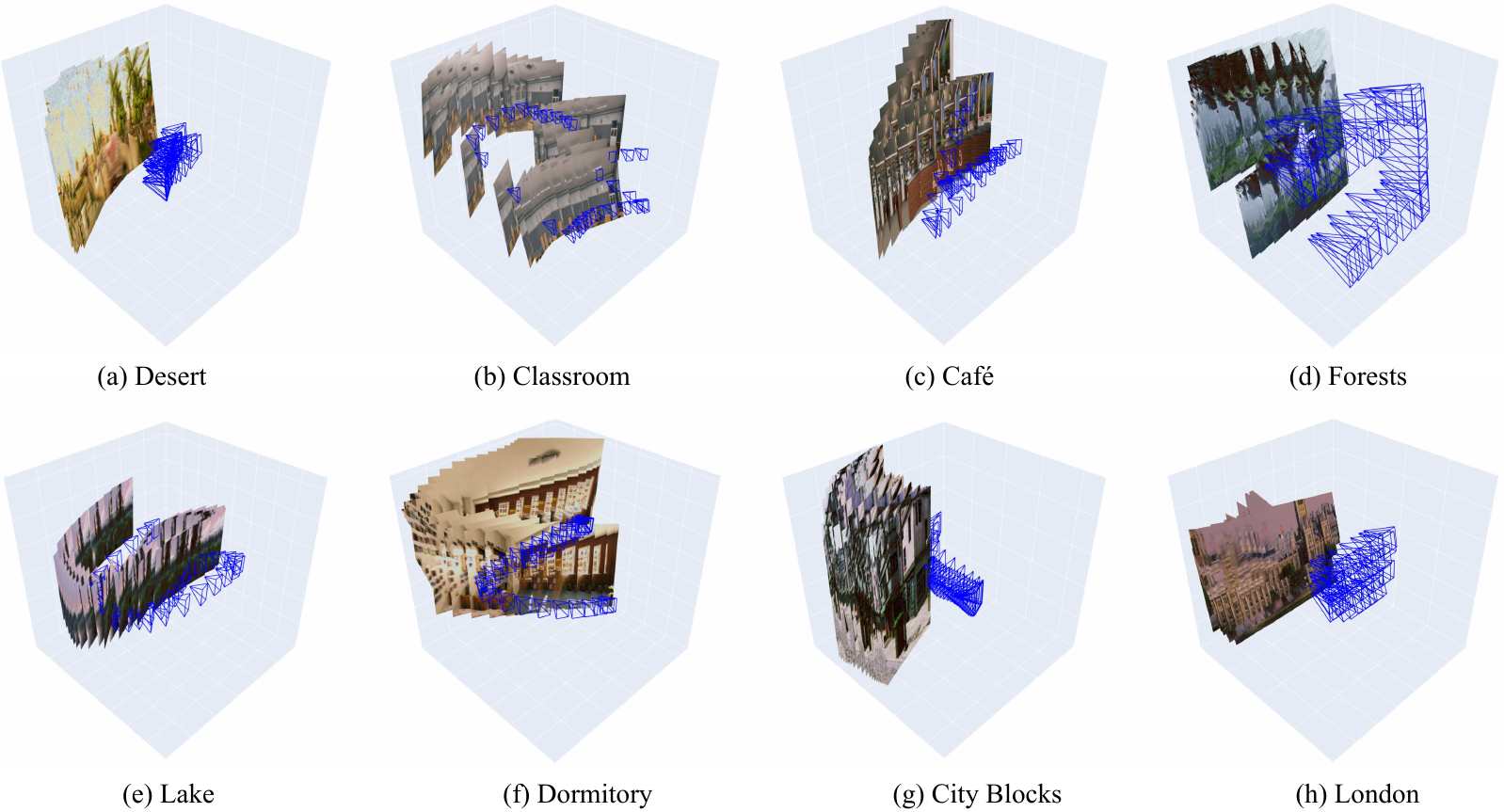}
    \end{center}
 \caption{Visualization of Camera Trajectory. The trajectories depicted were manually configured within Blender~\cite{Blender} to ensure precise control over the camera paths. For the purpose of visualization, these trajectories have been normalized.}
 \label{fig:traj}
\end{figure*}

\subsubsection{Camera Settings}
To render the base scenes and simulate motion blur, we configure the virtual camera in Blender with a resolution of \textnormal{$400\times600$}, and set the scaling factor to $1.0$. The virtual camera utilized a perspective model with a shutter speed of $1/180$ seconds. 
Subsequently, we developed a dedicated script to generate camera trajectory and motion blur. An example is shown in Figure.~\ref{fig:traj}. Along each predefined virtual camera trajectory, we uniformly sampled 35 camera poses, adding a certain level of jitter to create the training set. We recorded the start and end time of the camera exposure time, the positions, and 20 intermediate frames during the exposure time (obtained through linear interpolation between the start and end positions). We then uniformly sample 100 camera poses along the same trajectory to form the test set. Using the event camera simulator from V2E~\cite{vid2e}, we simulate the event stream for each camera trajectory and synthesize the event bins from the event stream at the start and end of the exposure time.

\subsection{EvaGaussians-DAVIS Dataset}
We use the color DAVIS346 event camera~\cite{davis346} to record our real-world event and RGB sequences and utilize the default camera settings provided in the DV software that comes with the camera. We name the five captured scenes as \textit{desk \& chair}, \textit{washroom}, \textit{pokémon}, \textit{pillow}, and \textit{bag}.

\subsubsection{Camera Calibration}
\begin{figure*}[h]
    \begin{center}
    \includegraphics[width=1\linewidth,trim={0.4cm 0.5cm 0.0cm 0.2cm},clip]{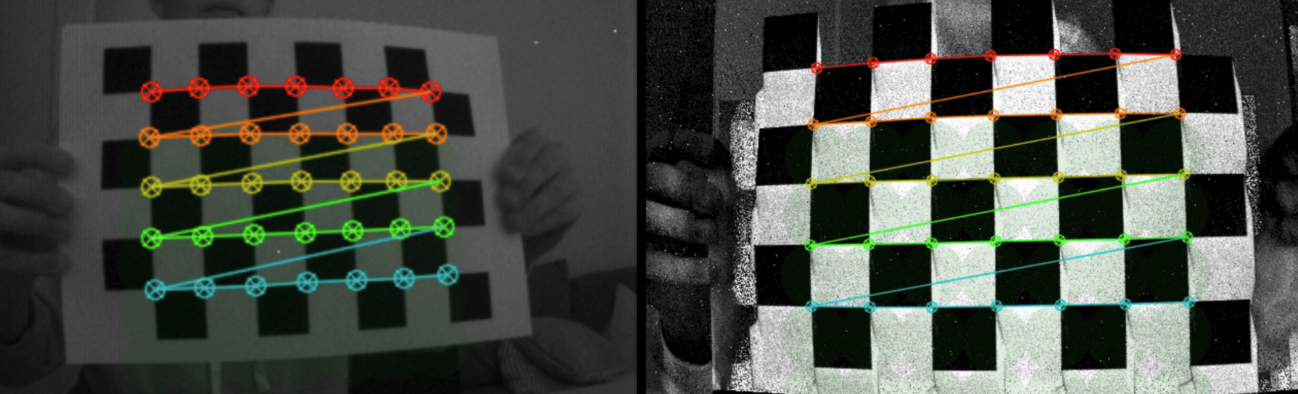}
    \end{center}
\caption{\textbf{Illustration of Camera Calibration.} The left panel shows the checkerboard pattern captured from various positions and angles, with detected corner points utilized for calibration. The right panel presents the calibrated checkerboard pattern, demonstrating the corresponding points and lines between two cameras, which reflect the geometric relationship and accuracy achieved after calibration. Different colored lines indicate the correspondences between points during the calibration process.}

    \label{fig:appendix_camera_calibration}
\end{figure*}

We calibrated the event camera using the DV software provided by DAVIS. During the calibration process, we used a \(6\times9\) checkerboard pattern with a square size of 30 mm. In the software configuration, we set the width to 9, height to 6, and square size to 30 mm. We then ran the calibration module and moved the calibration pattern in front of the camera. The software detected the pattern and collected images, highlighting the detected area in green, as shown in Figure.~\ref{fig:appendix_camera_calibration}. We set the minimum detections parameter to 50 to ensure a sufficient number of samples and used the consecutive detections parameter to ensure consistent pattern detection. Additionally, we enabled the image verification option to check the collected images in real-time, discarding inaccurately detected images and replacing them with new ones.
We evaluate the calibration accuracy using the reprojection error $e_r$ as Eq.~\ref{eq:reprojection_error} and the epipolar error $e_{\text{epipolar}}$ as Eq.~\ref{eq:epipolar_error} in stereo calibration. The reprojection error is calculated as follows:
\begin{equation}
e_r = \frac{1}{n} \sum_{i=1}^{n} \| x_i - \hat{x}_i \|
\label{eq:reprojection_error}
\end{equation}
where \( x_i \) represents the observed points and \( \hat{x}_i \) represents the projected points. The epipolar error is calculated as the average epipolar error for each point in all collected images. For each pair of images, the error is calculated as the sum of the distances between the points in one camera and the epipolar lines calculated from the other camera (\( m \) is the number of acquired images, \( n \) is the number of points). The formula is as follows:
\begin{equation}
e_{\text{epipolar}} = \frac{1}{m \times n} \sum_{i=1}^{m} \sum_{j=1}^{n} \left[ d(P1_{i,j}, l_{2,i,j}) + d(P2_{i,j}, l_{1,i,j}) \right]
\label{eq:epipolar_error}
\end{equation}
where \( P1_{i,j} \) and \( P2_{i,j} \) are the projection points of the \( j \)th point in the \( i \)th image in two cameras, and \( l_{1,i,j} \) and \( l_{2,i,j} \) are the epipolar lines corresponding to the \( j \)th point in the \( i \)th image calculated from the other camera. The maximum allowable error can be set under \textit{Max Reprojection Error}. The stereo calibration also calculates the error caused by the epipolar constraint, which can be set under \textit{Max Epipolar Error}. Once the calibration is successful, the results are saved and the undistorted output is displayed. The visualization results of the two types of errors are shown in Figure.~\ref{fig:appendix_error}. This process ensures the accuracy of the calibration, thereby improving the measurement accuracy and stability in subsequent applications.

\subsubsection{Camera Settings}
We recorded the five real scenes using the calibration parameters obtained during the calibration process. By adjusting the indoor lighting and shooting angles, we ensured the richness of the recorded scene details. 
The adopted event camera has a spatial resolution of $346\times260$, a temporal resolution of 1 \(\mu\)s, a typical latency of less than 1 ms, a maximum throughput of 12 MEps, and a dynamic range of approximately 120 $dB$ (with 50\% of the pixels responding to 80\% contrast changes under 0.1-100k lux conditions). The contrast sensitivity is 14.3\% (ON) and 22.5\% (OFF) (with 50\% of the pixels responding). These parameters ensure that the event camera can stably and efficiently record scene information under various lighting conditions and dynamic ranges.

\begin{table*}[h]
	\centering
        \caption{Quantitative comparisons of \textbf{DVS} on object-level scenes. The results indicate that our method outperforms previous state-of-the-art approaches, consistently achieving better performance across all metrics.}
    \label{tab:appendix_benchmark_on_e2nerf_dataset_per_scene_blur_view}
    \renewcommand{\arraystretch}{1.5}
    \resizebox{1.0\linewidth}{!}{
        \begin{tabular}{c||c c | c c c| c c c c||>{\centering\arraybackslash}p{1.35cm}}
            \toprule
                \textbf{Deblur View} & \textbf{{B-NeRF}} & \textbf{{{B-3DGS}}} & \textbf{UFP-GS} & \textbf{EDI-GS} & \textbf{EFN-GS} & \(\mathbf{E}^2\)\textbf{NeRF} & \textbf{BAD-NeRF} & \textbf{BAD-GS} & \textbf{EDNeRF} &  \textbf{Ours}\\
                \hhline{-||---------||-}
                {PSNR$\uparrow$}    & 22.87 & 23.01 & 27.99 & 27.92 & 28.12 & 29.70 & 28.33 & 28.61 & 29.95 & \textbf{30.02} \\
                {SSIM$\uparrow$}    & .9068 & .9092 & .9501 & .9495 & .9508 & .9589 & .9576 & .9582 & .9599 & \textbf{.9605} \\
                {LPIPS$\downarrow$} & .1450 & .1437 & .0743 & .0747 & .0739 & .0722 & .0734 & .0732 & .0720 & \textbf{.0719} \\
            \bottomrule
        \end{tabular}
    } 
\end{table*}

\begin{table*}[h]
	\centering
\caption{Quantitative comparisons of \textbf{DVS} on the medium-scale scenes. The results show that our method surpasses previous state-of-the-art approaches, achieving better performance consistently across all metrics.}
    \label{tab:appendix_benchmark_on_evags_deblur_view_medium_scale}
    \renewcommand{\arraystretch}{1.35}
	\resizebox{\linewidth}{!}{
        \begin{tabular}{c||c c | c c c| c c c c||>{\centering\arraybackslash}p{1.35cm}}
            \toprule
                \textbf{Deblur View} & \textbf{{B-NeRF}} & \textbf{{{B-3DGS}}} & \textbf{UFP-GS} & \textbf{EDI-GS} & \textbf{EFN-GS} & \(\mathbf{E}^2\)\textbf{NeRF} & \textbf{BAD-NeRF} & \textbf{BAD-GS} & \textbf{EDNeRF} &  \textbf{Ours}\\
                \hhline{-||---------||-}
                {PSNR$\uparrow$}    & 24.27 & 25.05 & 26.60 & 26.65 & 26.30 & 27.95 & 28.62 & 28.70 & 29.12 & \textbf{30.26} \\
                {SSIM$\uparrow$}    & .7254 & .7631 & .8135 & .8100 & .8068 & .8743 & .8883 & .8890 & .8951 & \textbf{.9241} \\
                {LPIPS$\downarrow$} & .3513 & .3101 & .2547 & .2486 & .2628 & .1874 & .1735 & .1715 & .1588 & \textbf{.1419} \\
            \bottomrule
        \end{tabular}
	} 
\end{table*}
\begin{table*}[!h]
	\centering
	\caption{Quantitative comparison of \textbf{DVS} on large-scale scenes. The results demonstrate that our method consistently achieves better performance across all metrics.}
    \label{tab:appendix_benchmark_on_evags_deblur_view}
    \renewcommand{\arraystretch}{1.35}
	\resizebox{\linewidth}{!}{
        \begin{tabular}{c||c c | c c c| c c c c||>{\centering\arraybackslash}p{1.35cm}}
            \toprule
                \textbf{Deblur View} & \textbf{{B-NeRF}} & \textbf{{{B-3DGS}}} & \textbf{UFP-GS} & \textbf{EDI-GS} & \textbf{EFN-GS} & \(\mathbf{E}^2\)\textbf{NeRF} & \textbf{BAD-NeRF} & \textbf{BAD-GS} & \textbf{EDNeRF} &  \textbf{Ours}\\
                \hhline{-||---------||-}
                {PSNR$\uparrow$}    & 21.42 & 21.58 & 21.43 & 22.36 & 22.75 & 23.02 & 23.92 & 23.98 & 24.79 & \textbf{25.85} \\
                {SSIM$\uparrow$}    & .6795 & .6914 & .6690 & .6943 & .6915 & .7155 & .7412 & .7425 & .7614 & \textbf{.8039} \\
                {LPIPS$\downarrow$} & .4185 & .3860 & .3672 & .3710 & .3520 & .3689 & .3468 & .3459 & .3168 & \textbf{.2635} \\
            \bottomrule
        \end{tabular}
	} 
\end{table*}

\begin{table*}[h]
    \caption{\textbf{The Novel View Synthesis Results of \colorbox{gray!15}{PSNR $\uparrow$} in the EvaGaussians-Blender Dataset}. The highest values in each category are highlighted in \textbf{bold} to indicate the best results.} 
    \label{tab:appendix_evags_dataset_psnr_nvs}
    \centering
    \setlength{\tabcolsep}{2pt}
    \resizebox{1.0\linewidth}{!}{%
        \begin{tabular}{lccccccccccccc}
        \toprule
        \multirow{2}{*}{\textbf{Models}} & \multicolumn{7}{c}{\textbf{Medium}} & \multicolumn{5}{c}{\textbf{Large}} \\
        \cmidrule(r){2-8}
        \cmidrule(r){9-13}
        & \textbf{Classroom} & \textbf{Dormitory} & \textbf{Café} & \textbf{Pool} & \textbf{Cozyroom} & \textbf{Factory} & \textbf{Tanabata} & \textbf{Desert} & \textbf{City Blocks} & \textbf{London} & \textbf{Forests} & \textbf{Lake}\\
        
        \midrule            
        Blurry-NeRF & 25.42 & 26.72 & 21.23 & 26.66 & 15.42 & 26.76 & 26.38 & 19.04 & 19.34 & 20.00 & 23.60 & 24.67 \\
        Blurry-GS & 25.59 & 26.83 & 21.34 & 26.17 & 20.76 & 26.48 & 26.46 & 19.67 & 19.48 & 20.18 & 23.75 & 24.34 \\
        \midrule  
        UFP-GS & 28.64 & 29.68 & 21.54 & 26.79 & 21.62 & 27.60 & 28.77 & 20.78 & 20.85 & 16.61 & 22.96 & 25.60 \\
        EDI-GS & 28.49 & 29.90 & 22.62 & 26.36 & 21.17 & 29.06 & 27.46 & 20.66 & 20.68 & 20.27 & 23.97 & 25.99 \\
        EFNET-GS & 28.59 & 29.48 & 22.04 & 26.36 & 21.42 & 27.57 & 27.43 & 20.62 & 21.70 & 20.36 & 25.57 & 25.18 \\
        \midrule  
        \( \textnormal{E}^2 \)\textnormal{NeRF} & 28.87 & 30.77 & 26.16 & 28.92 & 21.23 & 29.85 & 28.65 & 20.02 & 21.78 & 21.30 & 25.90 & 25.78 \\
        BAD-NeRF & 30.28 & 31.23 & 27.18 & 28.71 & 21.68 & 30.28 & 29.85 & 21.09 & 21.91 & 22.91 & 26.35 & 26.99 \\
        BAD-Gaussians & 30.28 & 31.23 & 27.18 & 28.72 & 21.68 & 30.28 & 29.85 & 21.10 & 21.93 & 22.93 & 26.35 & 26.98 \\
        EvDeblurNeRF & 31.83 & 28.95 & 28.66 & 29.69 & 22.01 & 31.20 & 30.02 & 21.62 & 22.23 & 23.88 & 27.10 & 28.29 \\
        \midrule  
        \textbf{Ours} & \textbf{34.38} & \textbf{32.97} & \textbf{30.41} & \textbf{30.26} & \textbf{22.71} & \textbf{31.85} & \textbf{30.71} & \textbf{24.88} & \textbf{23.71} & \textbf{23.99} & \textbf{27.62} & \textbf{29.90}  \\
        \bottomrule
    \end{tabular}
    }%
\end{table*}

\begin{table*}[h]
    \caption{\textbf{The Novel View Synthesis of \colorbox{gray!15}{SSIM $\uparrow$} in the EvaGaussians-Blender Dataset.} The highest values in each category are highlighted in \textbf{bold} to indicate the best results.} 
    \label{tab:appendix_evags_dataset_ssim_nvs}
    \centering
    \setlength{\tabcolsep}{2pt}
    \resizebox{1.0\linewidth}{!}{%
        \begin{tabular}{lccccccccccccc}
        \toprule
        \multirow{2}{*}{\textbf{Models}} & \multicolumn{7}{c}{\textbf{Medium}} & \multicolumn{5}{c}{\textbf{Large}} \\
        \cmidrule(r){2-8}
        \cmidrule(r){9-13}
        & \textbf{Classroom} & \textbf{Dormitory} & \textbf{Café} & \textbf{Pool} & \textbf{Cozyroom} & \textbf{Factory} & \textbf{Tanabata} & \textbf{Desert} & \textbf{City Blocks} & \textbf{London} & \textbf{Forests} & \textbf{Lake}\\

        \midrule            
        Blurry-NeRF & 0.7086 & 0.8281 & 0.5682 & 0.7442 & 0.5098 & 0.8567 & 0.8057 & 0.6023 & 0.6325 & 0.6732 & 0.7002 & 0.7823 \\
        Blurry-GS & 0.7154 & 0.8312 & 0.5638 & 0.7265 & 0.7632 & 0.8519 & 0.8064 & 0.6386 & 0.6341 & 0.6819 & 0.7026 & 0.7807 \\
        \midrule
        UFP-GS & 0.8701 & 0.9281 & 0.5706 & 0.7527 & 0.7729 & 0.8640 & 0.8569 & 0.6172 & 0.6144 & 0.5807 & 0.6911 & 0.7968 \\
        EDI-GS & 0.8456 & 0.9291 & 0.5977 & 0.7288 & 0.7703 & 0.8963 & 0.8408 & 0.6567 & 0.5655 & 0.6882 & 0.7158 & 0.8012 \\
        EFNET-GS & 0.8692 & 0.9259 & 0.5841 & 0.7288 & 0.7797 & 0.8643 & 0.8345 & 0.6164 & 0.6366 & 0.6069 & 0.7628 & 0.7901 \\
        \midrule
        \( \textnormal{E}^2 \)\textnormal{NeRF} & 0.8723 & 0.9319 & 0.7869 & 0.8795 & 0.7724 & 0.9245 & 0.8915 & 0.6169 & 0.6822 & 0.6734 & 0.7467 & 0.8139 \\
        BAD-NeRF & 0.8978 & 0.9337 & 0.8041 & 0.8794 & 0.7764 & 0.9353 & 0.9271 & 0.6314 & 0.6867 & 0.6932 & 0.7583 & 0.8919 \\
        BAD-Gaussians & 0.8992 & 0.9351 & 0.8031 & 0.8781 & 0.7755 & 0.9345 & 0.9266 & 0.6346 & 0.6851 & 0.6944 & 0.7578 & 0.8908 \\
        EvDeblurNeRF & 0.9023 & 0.8935 & 0.8513 & 0.8885 & 0.7854 & 0.9454 & 0.9311 & 0.6589 & 0.7023 & \textbf{0.7159} & 0.7726 & 0.9129 \\
        \midrule
        \textbf{Ours} & \textbf{0.9402} & \textbf{0.9580} & \textbf{0.9033} & \textbf{0.9108} & \textbf{0.8052} & \textbf{0.9592} & \textbf{0.9382} & \textbf{0.8152} & \textbf{0.7405} & 0.7145 & \textbf{0.8152} & \textbf{0.9465} \\
        \bottomrule
        
    \end{tabular}
    }%
\end{table*}

\begin{table*}[h]
    \caption{\textbf{The Novel View Synthesis of \colorbox{gray!15}{LPIPS $\downarrow$} in the EvaGaussians-Blender Dataset.} The highest values in each category are highlighted in \textbf{bold} to indicate the best results.} 
    \label{tab:appendix_evags_dataset_lpips_nvs}
    \centering
    \setlength{\tabcolsep}{2pt}
    \resizebox{1.0\linewidth}{!}{%
        \begin{tabular}{lcccccccccccc}
        \toprule
        \multirow{2}{*}{\textbf{Models}} & \multicolumn{7}{c}{\textbf{Medium}} & \multicolumn{5}{c}{\textbf{Large}} \\
        \cmidrule(r){2-8}
        \cmidrule(r){9-13}
        & \textbf{Classroom} & \textbf{Dormitory} & \textbf{Café} & \textbf{Pool} & \textbf{Cozyroom} & \textbf{Factory} & \textbf{Tanabata} & \textbf{Desert} & \textbf{City Blocks} & \textbf{London} & \textbf{Forests} & \textbf{Lake}\\
        
        \midrule            
        Blurry-NeRF & 0.3987 & 0.2998 & 0.4528 & 0.3848 & 0.5545 & 0.2063 & 0.2348 & 0.4447 & 0.5273 & 0.3971 & 0.3425 & 0.4127 \\
        Blurry-GS & 0.3824 & 0.2873 & 0.4554 & 0.4198 & 0.2432 & 0.2091 & 0.2335 & 0.4231 & 0.4116 & 0.3925 & 0.3379 & 0.4206 \\
        \midrule 
        UFP-GS & 0.2838 & 0.1361 & 0.4511 & 0.3646 & 0.2331 & 0.1928 & 0.1856 & 0.3816 & 0.3342 & 0.4069 & 0.3383 & 0.4069 \\
        EDI-GS & 0.2914 & 0.1267 & 0.4038 & 0.4102 & 0.2388 & 0.1446 & 0.1911 & 0.4062 & 0.3857 & 0.3901 & 0.3295 & 0.4001 \\
        EFNET-GS & 0.2846 & 0.1373 & 0.4529 & 0.4102 & 0.2345 & 0.1926 & 0.1958 & 0.3912 & 0.3222 & 0.3932 & 0.2962 & 0.4128 \\
        \midrule 
        \( \textnormal{E}^2 \)\textnormal{NeRF} & 0.2821 & 0.1165 & 0.2871 & 0.2048 & 0.2369 & 0.1073 & 0.1546 & 0.3938 & 0.3713 & 0.3928 & 0.3589 & 0.3588 \\
        BAD-NeRF & 0.2365 & 0.1083 & 0.2715 & 0.2103 & 0.2278 & 0.0992 & 0.1224 & 0.3947 & 0.3618 & 0.3224 & 0.3467 & 0.3142 \\
        BAD-Gaussians & 0.2384 & 0.1078 & 0.2695 & 0.2094 & 0.2262 & 0.0985 & 0.1215 & 0.3965 & 0.3604 & 0.3215 & 0.3452 & 0.3129 \\
        EvDeblurNeRF & 0.2217 & 0.1455 & 0.2447 & 0.1926 & 0.1942 & 0.0768 & 0.1086 & 0.3823 & 0.3497 & \textbf{0.2972} & 0.3163 & 0.2941 \\
        \midrule 
        \textbf{Ours} & \textbf{0.1927} & \textbf{0.0928} & \textbf{0.2311} & \textbf{0.1859} & \textbf{0.1873} & \textbf{0.0715} & \textbf{0.1023} & \textbf{0.2053} & \textbf{0.2835} & 0.2983 & \textbf{0.2857} & \textbf{0.2674} \\

        \bottomrule
        
    \end{tabular}
    }%
\end{table*}

\begin{table*}[h]
    \caption{The \textbf{Novel View Synthesis} of \colorbox{gray!15}{BRISQUE} in the \textbf{EvaGaussians-DAVIS} Dataset. The highest values in each category are highlighted in \textbf{bold} to indicate the best results.} 
    \label{tab:appendix_evags_dataset_brisque_nvs}
    \centering
     \setlength{\tabcolsep}{17pt}
    \resizebox{1.\linewidth}{!}{%
        \begin{tabular}{lcccccc}
        \toprule
        \multirow{2}{*}{\textbf{Models}} & \multicolumn{5}{c}{\textbf{BRISQUE} $\downarrow$}\\
        \cmidrule(r){2-7}
                                        & \textbf{Desk \& Chair} & \textbf{Washroom} & \textbf{Pokémon} & \textbf{Pillow} & \textbf{Bag} & \textbf{Average} \\
        \midrule            
        B-NeRF & 63.9428 & 102.7828 & 109.2711 & 97.4778 & 87.7699 & 92.2489 \\
        B-3DGS & 51.1542 & 82.2262 & 87.4169 & 77.9823 & 70.2159 & 73.7991 \\
        UFP-GS & 43.6932 & 69.7354 & 74.6684 & 66.6098 & 59.9711 & 62.9356 \\
        EDI-GS & 43.4811 & 69.9923 & 74.3044 & 66.2849 & 59.6835 & 62.7492 \\
        EFN-GS & 43.5235 & 69.8473 & 74.2875 & 66.3158 & 59.6523 & 62.7253 \\
        \midrule
        \({\textnormal{E}^2}\)\textnormal{NeRF} & 36.2148 & 64.4332 & 77.0112 & 67.1324 & 62.8063 & 61.5196 \\
        BAD-NeRF & 42.6285 & 68.5219 & 72.8474 & 64.9852 & 58.5133 & 61.4993 \\
        BAD-GS & 42.2065 & 67.8434 & 72.1261 & 64.3418 & 57.9339 & 60.8903 \\
        EDNeRF & 34.5687 & 61.5044 & 73.5012 & 64.0809 & 59.4969 & 58.6304 \\
        \midrule
        \textbf{Ours} & \textbf{32.9225} & \textbf{58.5756} & \textbf{70.0211} & \textbf{61.0294} & \textbf{58.1876} & \textbf{56.1472}  \\
        \bottomrule
    \end{tabular}
    }%
\end{table*}

\begin{table*}[h]
\centering
 \setlength{\tabcolsep}{36pt}
\caption{Robustness of pose optimization.}
\label{tab:pose}
\resizebox{1.\linewidth}{!}{ 
\begin{tabular}{c|c|c|c}
\toprule
\textbf{ATE} $\downarrow$ & \textbf{Mild Blur} & \textbf{Medium Blur}  & \textbf{Strong Blur}\\
\midrule
\textbf{COLMAP poses} & $0.0598 \pm 0.0167$ & $0.0783 \pm 0.0245$ & $0.0973 \pm 0.0352$\\
\textbf{Optimized poses}  & $0.0461 \pm 0.0129$ & $0.0586 \pm 0.0149$& $0.0677 \pm 0.0208$\\
\bottomrule
\end{tabular}
}
\end{table*}

\begin{table*}[h]
    \caption{The \textbf{Novel View Synthesis} of \colorbox{gray!15}{NIQE} in the \textbf{EvaGaussians-DAVIS} Dataset.} 
    \label{tab:appendix_evags_dataset_niqe_nvs}
    \centering
    \setlength{\tabcolsep}{17pt}
    \resizebox{1.\linewidth}{!}{%
        \begin{tabular}{lcccccc}
        \toprule
        \multirow{2}{*}{\textbf{Models}} & \multicolumn{5}{c}{\textbf{NIQE} $\downarrow$}  \\
        \cmidrule(r){2-7}
                                        & \textbf{Desk \& Chair} & \textbf{Washroom} & \textbf{Pokémon} & \textbf{Pillow} & \textbf{Bag}& \textbf{Average}\\
        \midrule            
        B-NeRF & 11.4380 & 15.5386 & 16.9153 & 16.2316 & 14.8844 & 15.0016 \\
        B-3DGS & 9.1504 & 12.4309 & 13.5323 & 12.9853 & 11.9075 & 12.0113 \\
        UFP-GS & 7.7736 & 10.6170 & 11.3827 & 10.9266 & 10.1715 & 10.1743 \\
        EDI-GS & 7.7778 & 10.5662 & 11.5024 & 11.0375 & 10.1214 & 10.2011 \\
        EFN-GS & 7.8235 & 10.5412 & 11.4868 & 11.0743 & 10.1288 & 10.2109 \\
        \midrule
        \({\textnormal{E}^2}\)\textnormal{NeRF} & 6.8907 & 9.0924 & 11.6383 & 10.1830 & 9.3954 & 9.4400 \\
        BAD-NeRF & 7.6253 & 10.3590 & 11.2769 & 10.8211 & 9.9229 & 10.0011 \\
        BAD-GS & 7.5498 & 10.2565 & 11.1652 & 10.7139 & 9.8247 & 9.9020 \\
        EDNeRF & 6.5775 & 8.6791 & 11.1092 & 9.7201 & 8.9684 & 9.0109 \\
        \midrule
                \textbf{Ours} & \textbf{6.1898} & \textbf{8.1117} & \textbf{10.2662} & \textbf{8.9455} & \textbf{8.3421} & \textbf{8.3711}  \\
        \bottomrule
    \end{tabular}
    }%
\end{table*}

\begin{table*}[h]
    \caption{The \textbf{Novel View Synthesis} of \colorbox{gray!15}{PIQE} in the \textbf{EvaGaussians-DAVIS} Dataset.} 
    \label{tab:appendix_evags_dataset_piqe_nvs}
    \centering
    \setlength{\tabcolsep}{17pt}
    \resizebox{1.\linewidth}{!}{%
        \begin{tabular}{lcccccc}
        \toprule
        \multirow{2}{*}{\textbf{Models}} & \multicolumn{5}{c}{\textbf{PIQE} $\downarrow$}  \\
        \cmidrule(r){2-7}
                                        & \textbf{Desk \& Chair} & \textbf{Washroom} & \textbf{Pokémon} & \textbf{Pillow} & \textbf{Bag}& \textbf{Average}\\
        \midrule            
        B-NeRF & 58.4566 & 68.8701 & 73.4701 & 54.0862 & 74.7257 & 65.9217 \\
        B-3DGS & 46.7653 & 55.0961 & 58.7761 & 43.2689 & 59.7806 & 52.7374 \\
        UFP-GS & 39.9453 & 47.0612 & 50.2026 & 36.9659 & 50.9631 & 45.0276 \\
        EDI-GS & 39.7505 & 46.8317 & 49.9597 & 36.7786 & 50.8135 & 44.8268 \\
        EFN-GS & 39.7748 & 46.7951 & 49.9346 & 36.7891 & 50.8275 & 44.8242 \\
        \midrule
        \({\textnormal{E}^2}\)\textnormal{NeRF} & 40.9825 & 49.2988 & 50.4868 & 39.7896 & 53.2203 & 46.7556 \\
        BAD-NeRF & 38.9711 & 45.9134 & 48.9801 & 36.0574 & 49.8171 & 43.9478 \\
        BAD-GS & 38.5852 & 45.4588 & 48.4951 & 35.7004 & 49.3239 & 43.5127 \\
        EDNeRF & 39.1197 & 47.0580 & 48.1919 & 37.9810 & 50.8012 & 44.6304 \\
        \midrule
        \textbf{Ours} & \textbf{36.5717} & \textbf{44.3304} & \textbf{44.0641} & \textbf{35.3059} & \textbf{47.3749} & \textbf{41.5294}  \\
        \bottomrule
    \end{tabular}
    }%
\end{table*}

\begin{table*}
    \caption{The \textbf{Novel View Synthesis} of \colorbox{gray!15}{MetaIQA} in the \textbf{EvaGaussians-DAVIS} Dataset.} 
    \label{tab:appendix_evags_dataset_metaiqa_nvs}
    \centering
    \setlength{\tabcolsep}{17pt}
    \resizebox{1.\linewidth}{!}{%
        \begin{tabular}{lcccccc}
        \toprule
        \multirow{2}{*}{\textbf{Models}} & \multicolumn{5}{c}{\textbf{MetaIQA} $\uparrow$}  \\
        \cmidrule(r){2-7}
                                        & \textbf{Desk \& Chair} & \textbf{Washroom} & \textbf{Pokémon} & \textbf{Pillow} & \textbf{Bag}& \textbf{Average}\\
        \midrule            
        B-NeRF & 0.1419 & 0.1272 & 0.1085 & 0.1122 & 0.1307 & 0.1241 \\
        B-3DGS & 0.1621 & 0.1454 & 0.1240 & 0.1283 & 0.1494 & 0.1418 \\
        UFP-GS & 0.1976 & 0.1774 & 0.1521 & 0.1564 & 0.1823 & 0.1732 \\
        EDI-GS & 0.1986 & 0.1780 & 0.1518 & 0.1571 & 0.1830 & 0.1737 \\
        EFN-GS & 0.1928 & 0.1835 & 0.1579 & 0.1643 & 0.1799 & 0.1757 \\
        \midrule
        \({\textnormal{E}^2}\)\textnormal{NeRF} & 0.1959 & 0.2020 & 0.1619 & 0.1723 & 0.1722 & 0.1809 \\
        BAD-NeRF & 0.2027 & 0.1817 & 0.1549 & 0.1603 & 0.1867 & 0.1773 \\
        BAD-GS & 0.2047 & 0.1835 & 0.1565 & 0.1620 & 0.1886 & 0.1790 \\
        EDNeRF & 0.2067 & 0.2132 & 0.1709 & 0.1819 & 0.1817 & 0.1909 \\
        \midrule
        \textbf{Ours} & \textbf{0.2135} & \textbf{0.2188} & \textbf{0.1975} & \textbf{0.2188} & \textbf{0.2160} & \textbf{0.2129}  \\
        \bottomrule
    \end{tabular}
    }%
\end{table*}

\begin{table*}
    \caption{The \textbf{Novel View Synthesis} of \colorbox{gray!15}{RankIQA} in the \textbf{EvaGaussians-DAVIS} Dataset.} 
    \label{tab:appendix_evags_dataset_rankiqa_nvs}
    \centering
    \setlength{\tabcolsep}{17pt}
    \resizebox{1.\linewidth}{!}{%
        \begin{tabular}{lcccccc}
        \toprule
        \multirow{2}{*}{\textbf{Models}} & \multicolumn{5}{c}{\textbf{RankIQA} $\downarrow$}  \\
        \cmidrule(r){2-7}
                                        & \textbf{Desk \& Chair} & \textbf{Washroom} & \textbf{Pokémon} & \textbf{Pillow} & \textbf{Bag}& \textbf{Average}\\
        \midrule            
        B-NeRF & 7.4896 & 10.4454 & 10.7921 & 9.6578 & 8.7541 & 9.4278 \\
        B-3DGS & 5.9917 & 8.3563 & 8.6337 & 7.7262 & 7.0032 & 7.5422 \\
        UFP-GS & 5.1153 & 7.1215 & 7.3741 & 6.6005 & 5.9829 & 6.4389 \\
        EDI-GS & 5.0929 & 7.1029 & 7.3386 & 6.5673 & 5.9528 & 6.4109 \\
        EFN-GS & 5.1046 & 7.0893 & 7.3158 & 6.5789 & 5.9682 & 6.4114 \\
        \midrule
        \({\textnormal{E}^2}\)\textnormal{NeRF} & 4.5141 & 6.4997 & 5.7764 & 5.7600 & 5.3166 & 5.5733 \\
        BAD-NeRF & 4.9931 & 6.9636 & 7.1948 & 6.4385 & 5.8360 & 6.2852 \\
        BAD-GS & 4.9436 & 6.8947 & 7.1235 & 6.3748 & 5.7783 & 6.2230 \\
        EDNeRF & 4.3089 & 6.2043 & 5.5138 & 5.4981 & 5.0749 & 5.3200 \\
        \midrule
        \textbf{Ours} & \textbf{3.9369} & \textbf{5.6564} & \textbf{5.0867} & \textbf{5.1538} & \textbf{4.6422} & \textbf{4.8952}  \\
        \bottomrule
    \end{tabular}
    }%
\end{table*}

\begin{figure*}
    \begin{center}
    \includegraphics[width=\linewidth]{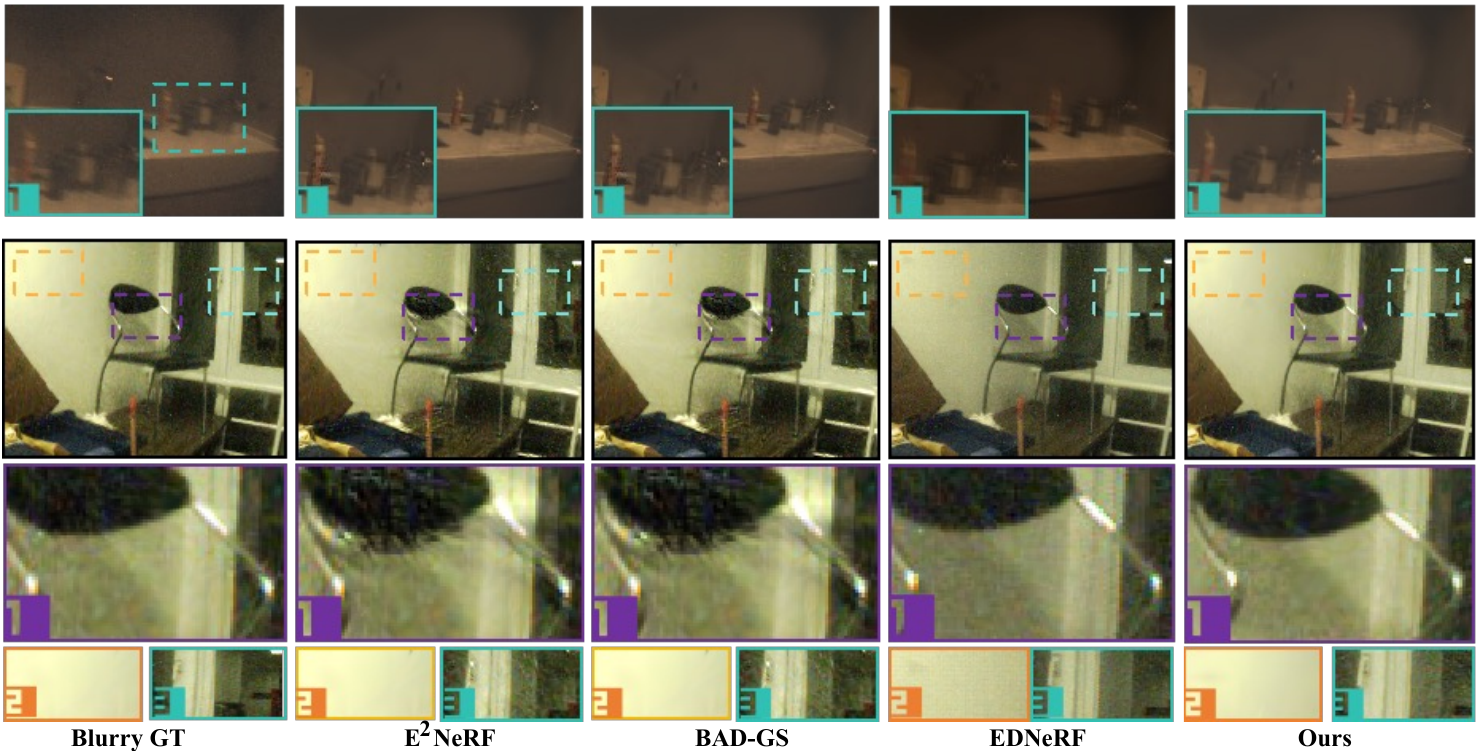}

    \end{center}
        \caption{NVS results on the \textbf{EvaGaussian-DAVIS} dataset. The first column shows the blurry image used for training, and the following rows show the deblurring results of different methods. The results demonstrate that our method consistently excels in reconstructing fine details compared to other methods~\cite{qi2023e2nerf,zhao2024badgs,evdeblurnerf}.}
    \label{fig:comp_real}
\end{figure*}

\begin{figure*}
    \begin{center}
    \includegraphics[width=\linewidth,trim={0.4cm 0.5cm 0.0cm 0.2cm},clip]{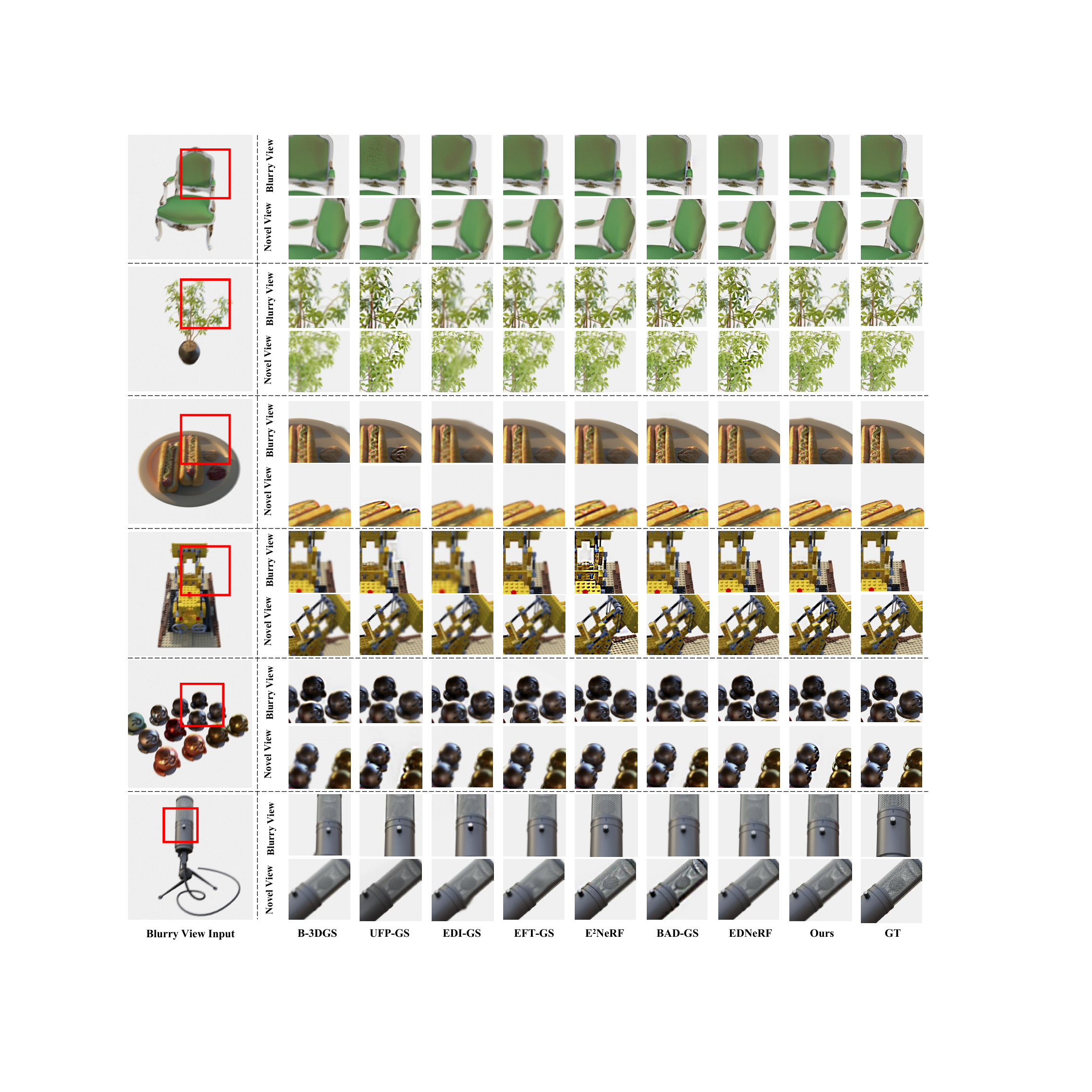}

    \end{center}
        \caption{\textbf{Visualization of DVS and NVS of Object-level Scenes in the EvaGaussian-Blender Dataset.} The DVS results are highlighted in the red bouding box. The results demonstrate that our method consistently excels in reconstructing fine details and maintaining high color accuracy compared to other methods~\cite{nerf,kerbl3Dgaussians,pan2019edi,efnet,qi2023e2nerf,zhao2024badgs,evdeblurnerf,ufp}.}
    \label{fig:appendix_vis_object_dvs_nvs}
\end{figure*}

\begin{figure*}
    \begin{center}
    \includegraphics[width=1\linewidth,trim={0.4cm 0.5cm 0.0cm 0.2cm},clip]{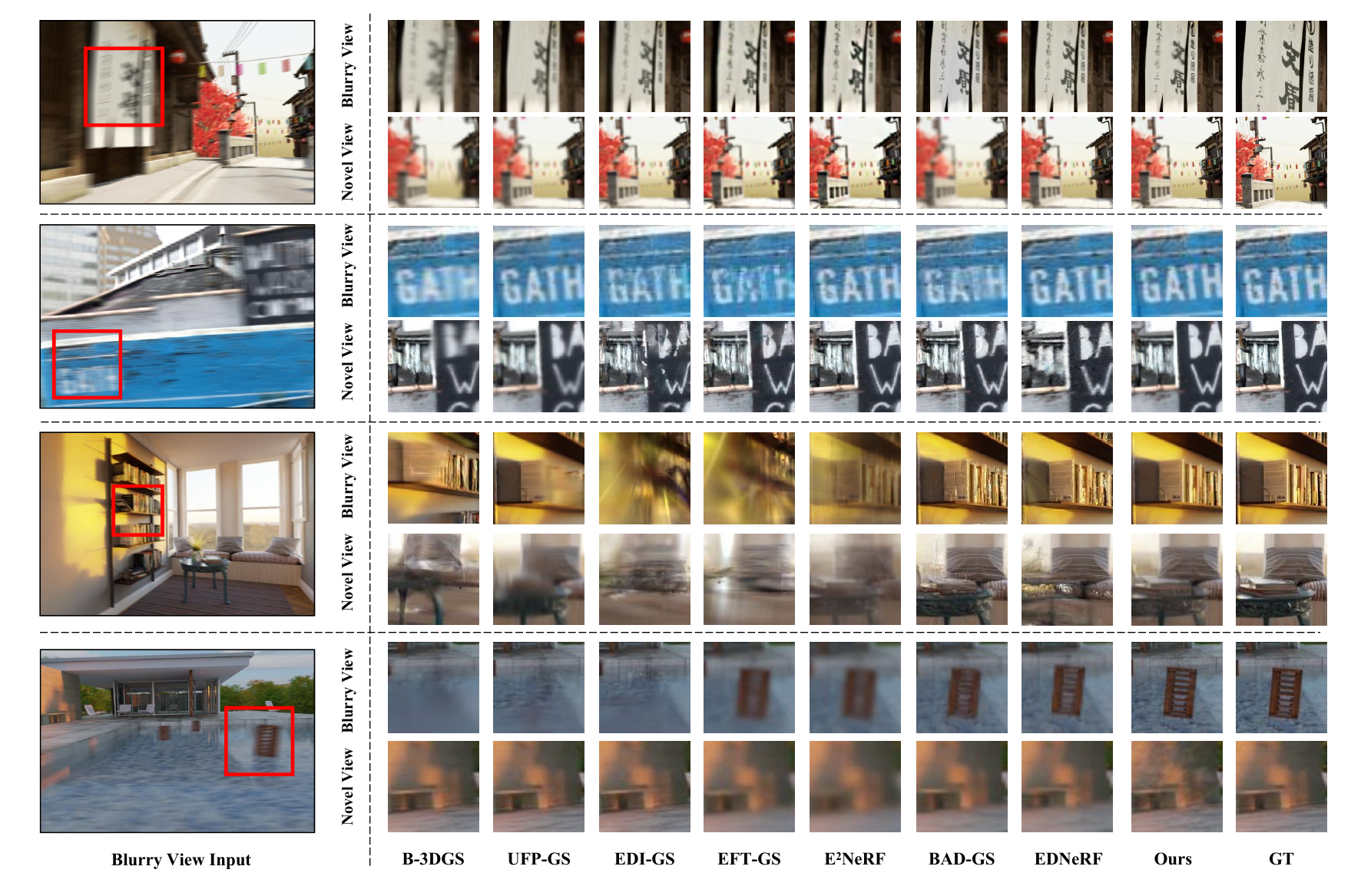}
    \end{center}
    \caption{\textbf{Visualization of DVS and NVS of results.} The DVS results are highlighted in the red bouding box. The results demonstrate that our method consistently excels in reconstructing fine details and maintaining high color accuracy compared to other methods.}
    \label{fig:appendix_deblurnerf_synthesis_dvs_nvs}
\end{figure*}

\begin{figure*}
    \begin{center}
    \includegraphics[width=.8\linewidth]{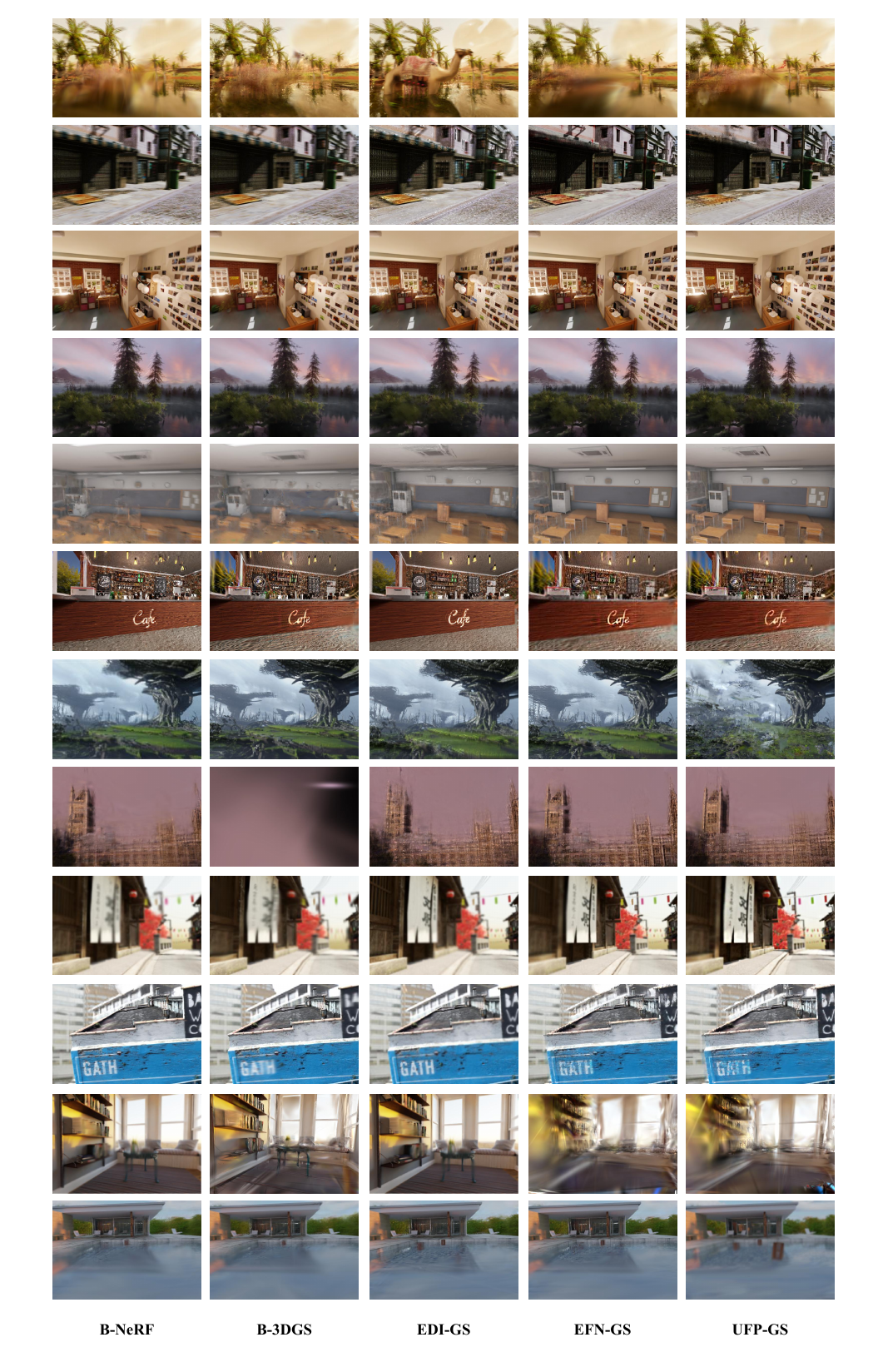}
    \end{center}
    \caption{\textbf{Visualization of Novel View Synthesis of All-redesigned Scenes with B-NeRF, B-3DGS, EDI-GS, EFN-GS and UFP-GS in the EvaGaussian-Blender Dataset.} }
    \label{fig:appendix_evags_nvs_1}
\end{figure*}

\begin{figure*}
    \begin{center}
    \includegraphics[width=.8\linewidth]{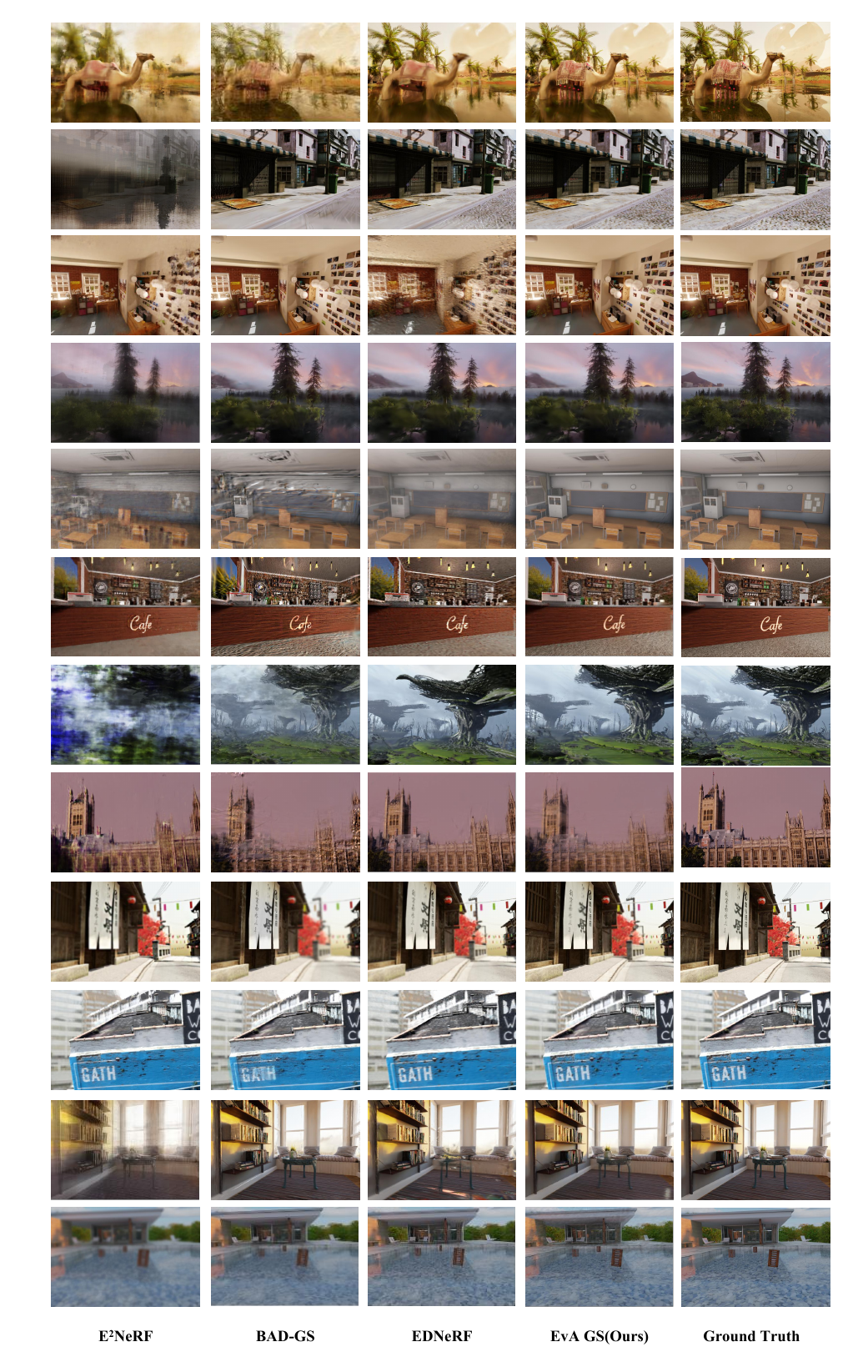}
    \end{center}
    \caption{\textbf{Visualization of Novel View Synthesis of All-redesigned Scenes with \({\textnormal{E}^2}\)\textnormal{NeRF}, BAD-GS, EDNeRF and EvAGS in the EvaGaussian-Blender Dataset.}}
    \label{fig:appendix_evags_nvs_2}
\end{figure*}

\section{Detailed Experiments}
\subsection{Synthetic Data Experiments}

\subsubsection{Deblurring View Synthesis Comparison}
\label{app:dvs_obj_scene}
Refering to~\cite{wang2023badnerf}, we additionally provide deblurring view synthesis (\textbf{DVS}) results on our proposed \textbf{EvaGaussians-Blender} dataset, and show more qualitative results of novel view synthesis (\textbf{NVS}).
For object-level scenes,
Table.~\ref{tab:appendix_benchmark_on_e2nerf_dataset_per_scene_blur_view} and Figure.~\ref{fig:appendix_vis_object_dvs_nvs} present the quantitative and qualitative results of ours and the comparison baselines across six synthetic scene sequences. From the qualitative results, it is evident that our method excels in reconstructing fine details and maintaining high fidelity in both NVS and DVS. In terms of quantitative results, our method outperforms baseline methods in most scenes. 
For medium-scale scenes and large-scale scenes, the quantitative results are shown in Table.~\ref{tab:appendix_benchmark_on_evags_deblur_view_medium_scale} and Table.~\ref{tab:appendix_benchmark_on_evags_deblur_view}, and the qualitative results of NVS and DVS are shown in  Figure.~\ref{fig:appendix_deblurnerf_synthesis_dvs_nvs}, which demonstrate that our model achieves better performance in both tasks.

\subsubsection{Per-scene Comparison for Novel View Synthesis}
\label{app:per-scene_evags_dataset}
In this subsection, we present a detailed per-scene analysis of the novel view synthesis performance in medium and large scale scenes from the \textbf{EvaGaussians-Blender} dataset, to evaluate the effectiveness of our method across different challenging scenes.

Table.~\ref{tab:appendix_evags_dataset_psnr_nvs} shows the PSNR value of the NVS results, which demonstrates that our proposed method consistently outperforms other approaches across various scenes. 
The detailed metrics for SSIM and LPIPS in Table.~\ref{tab:appendix_evags_dataset_ssim_nvs} and Table.~\ref{tab:appendix_evags_dataset_lpips_nvs}  further show that our model excels in maintaining structural integrity and perceptual quality in synthesized views.
Specifically, in medium-scale scenes, our method exhibits robust performance, particularly in complex environments where maintaining detail and minimizing artifacts are challenging. This can be demonstrated in scenes such as \textit{cozyroom} and \textit{factory}, where our method achieves significant improvements in both PSNR and SSIM.
For large-scale scenes, scenes like \textit{desert} and \textit{city blocks} highlight the model's capability to generalize across different scales and provide high-quality novel view synthesis.

We also present more qualitative results of novel view synthesis in Figure.~\ref{fig:appendix_evags_nvs_1} and Figure.~\ref{fig:appendix_evags_nvs_2}.
The results highlight our model's ability to reconstruct fine details and maintain high color accuracy beyond the comparison baselines. 

\subsection{Real-world Data Experiments}
\label{app:evags_davis_nvs_dvs}
In this section, we present a comprehensive per-scene analysis of \textbf{NVS} results on the \textbf{EvaGaussians-DAVIS} dataset. The qualitative results are shown in Figure.~\ref{fig:comp_real}, where the first column shows the blurry image used for training, and the following rows show the deblur results of different methods. The results demonstrate that our method consistently excels in reconstructing fine details compared to other methods.

We further report the per-scene quantitative results to validate our robustness to different scenes. As introudced in the main text, the adopted metrics include BRISQUE, NIQE, PIQE, MetaIQA, and RankIQA, which can effectively assess the quality of synthesized views in a no-reference manner.
As shown in Table.~\ref{tab:appendix_evags_dataset_brisque_nvs}, our model achieves the best BRISQUE scores across all scenes, highlighting its ability to produce visually appealing and less distorted images.
For NIQE, as presented in Table.~\ref{tab:appendix_evags_dataset_niqe_nvs}, our approach significantly outperforms the baselines, achieving the lowest average NIQE score. This demonstrates our method's robustness in generating high-quality images with minimal perceptual artifacts.
In terms of PIQE, Table.~\ref{tab:appendix_evags_dataset_piqe_nvs} shows that our model again leads in performance, achieving the lowest PIQE scores, which underscores the effectiveness of our model in preserving image details and reducing noise.
Furthermore, our method excels in MetaIQA and RankIQA evaluations, as detailed in Tables.~\ref{tab:appendix_evags_dataset_metaiqa_nvs} and Table.~\ref{tab:appendix_evags_dataset_rankiqa_nvs}, respectively. The highest MetaIQA scores and lowest RankIQA scores across most scenes affirm the overall better visual quality and fidelity of our synthesized views compared to baseline models.
Overall, these results demonstrate the robustness of our method, particularly in handling complex scenes and maintaining high visual quality across diverse scenarios. 

\subsection{Ablation Study}
In this section, we present an additional ablation study on the robustness of pose optimization in different blur level. We redesign three different levels of motion blur sequences in medium-scale scenes and compare the Average Trajectory Error (ATE) between the initial poses produced by COLMAP and the optimized poses. 
Figure.~\ref{fig:pose} illustrates the visualization of COLMAP poses and the optimized poses on the \textit{City Blocks} scene. As the motion blur becomes more severe, the accuracy of the COLMAP poses is significantly impacted, while the optimized poses maintain a higher level of accuracy. In a horizontal comparison, the optimized poses better match the ground truth across various levels of blur, demonstrating the effectiveness of pose optimization. 
Table.~\ref{tab:pose} presents the quantitative results, further demonstrating the effectiveness and robustness of pose optimization in handling different levels of motion blur. 
\section{Broader Impacts}
\label{app:broader}
Our proposed \textbf{EvaGaussians} leverages event cameras to assist novel view synthesis from low-quality, blurred images. It has the potential to bring about both positive and negative societal impacts. 

On the positive side, our method can improve the efficiency of surveillance systems by reconstructing clear 3D images from low-quality footage, enabling better identification of individuals and objects in challenging conditions. This can bolster public safety and aid in criminal investigations. Additionally, the ability to reconstruct scenes from blurred inputs can enhance the performance of autonomous vehicles, drones, and robots, enabling them to navigate more accurately in poor visibility conditions, leading to safer and more efficient transportation and logistics. In situations where traditional cameras may struggle to capture clear images under extreme conditions, our method can provide valuable information for first responders and rescue teams, helping them make informed decisions and potentially saving lives. Furthermore, our technique can be applied to medical imaging, allowing for better visualization of internal structures and more accurate diagnoses, ultimately leading to improved patient outcomes.

On the negative side, the enhanced surveillance capabilities enabled by our method may raise privacy concerns. For example, our method could be used for malicious purposes, such as stalking or spying on individuals without their consent. It is important to establish regulations and guidelines to prevent such misuse.

\end{document}